\documentclass{article}

\usepackage{iclr2027_conference,times}

\usepackage{amsmath,amsfonts,bm}

\def\eqref#1{equation~\ref{#1}}

\def\1{\bm{1}}

\DeclareMathAlphabet{\mathsfit}{\encodingdefault}{\sfdefault}{m}{sl}
\SetMathAlphabet{\mathsfit}{bold}{\encodingdefault}{\sfdefault}{bx}{n}

\usepackage{amsmath,amssymb}
\usepackage{dsfont}
\usepackage{booktabs}
\usepackage{array,longtable}
\usepackage{graphicx}
\ifdefined\pdfinclusioncopyfonts\pdfinclusioncopyfonts=1\fi
\usepackage{xcolor}
\usepackage{listings}
\usepackage{hyperref}
\hypersetup{
  pdftitle={Do LMs Know What Variables Do? Interpreting Code Representations via Variable Roles},
  pdfauthor={Célian Vasson, Daniil Ognev, Kentaro Inui, Benjamin Heinzerling}
}
\usepackage{url}
\usepackage{subcaption}
\usepackage{etoc}

\newif\ifnewdatasetappendices
\newdatasetappendicesfalse

\title{Do LMs Know What Variables Do? Interpreting Code Representations via Variable Roles}

\author{Célian Vasson$^{1}$ \quad Daniil Ognev$^{2}$\\[0.35em]
\bfseries Kentaro Inui$^{2,3,4}$ \quad Benjamin Heinzerling$^{3,4}$\\[0.8em]
\normalfont\small $^{1}$Sorbonne Université\\
$^{2}$MBZUAI\\
$^{3}$RIKEN AIP\quad $^{4}$Tohoku University}

\iclrfinalcopy

\begin{document}
\etocdepthtag.toc{main}

\maketitle
\fancyhead[L]{}

\begin{abstract}
Variable roles describe recurring behaviors such as accumulation, traversal, and retaining the best candidate encountered.
We use Sajaniemi's independently defined taxonomy to study which role distinctions can be recovered from large language models, how sparse autoencoders (SAEs) retain those distinctions, and whether associated directions support control of generation.
We introduce static annotation rules for all eleven roles and a corpus of 126,926 natural-code files across five languages.
Across five model--SAE systems, residual probes outperform individual SAE features and tested lexical baselines, with modest losses after neutral target renaming.
Combining SAE features closes much of the single-feature gap, but probes on SAE reconstructions are less sensitive to role-changing edits reviewed by LLMs.
High paired ranking accuracy coexists with frequent false positives at fixed thresholds.
During generation, interventions often change role annotations at the cost of correctness; changes accompanied by passing tests show no clear aggregate advantage over random controls.
The benchmark separates role-related recoverability, SAE fidelity, and correctness-preserving control through matched prediction, source edits, and generation interventions.
\end{abstract}

\section{Introduction}
Large language models generate and modify source code, from code completion to agents that plan, edit, and test software \citep{jiang2024survey,dong2025agents}.
Yet programming success alone does not establish which program properties a model represents: models may exploit identifiers and familiar code patterns, and their performance can change under controlled input transformations \citep{hooda2024cacp}.
We investigate this question through the behavior of individual variables.
A variable may advance through a predictable sequence, accumulate values, or retain the best candidate seen so far.
These behaviors depend on relationships across updates: recognizing an addition alone does not distinguish a regular progression from accumulation of input-dependent data.
The same behavior can also use different names and language constructs, motivating three questions: \emph{which role distinctions are recoverable, how faithfully do sparse autoencoders retain them, and can associated directions change generation while preserving correctness?}

We investigate these questions using Sajaniemi's \emph{variable roles}, developed in the psychology of programming to characterize recurring uses of variables \citep{sajaniemi2002visualizing,sajaniemi2005experts}.
The taxonomy defines behavioral criteria independently of the models under study.
Prior studies recover program structure and intermediate execution states from model representations \citep{hernandezlopez2022astprobe,jin2024semantics} and find shared representations across programming languages \citep{kargaran2025shared,yin2026neuronguided}.
We extend these analyses to variable behavior through matched prediction, controlled source edits, and generation interventions.

\begin{figure}[t]
    \centering
    \setlength{\abovecaptionskip}{6pt}
    \includegraphics[width=\linewidth]{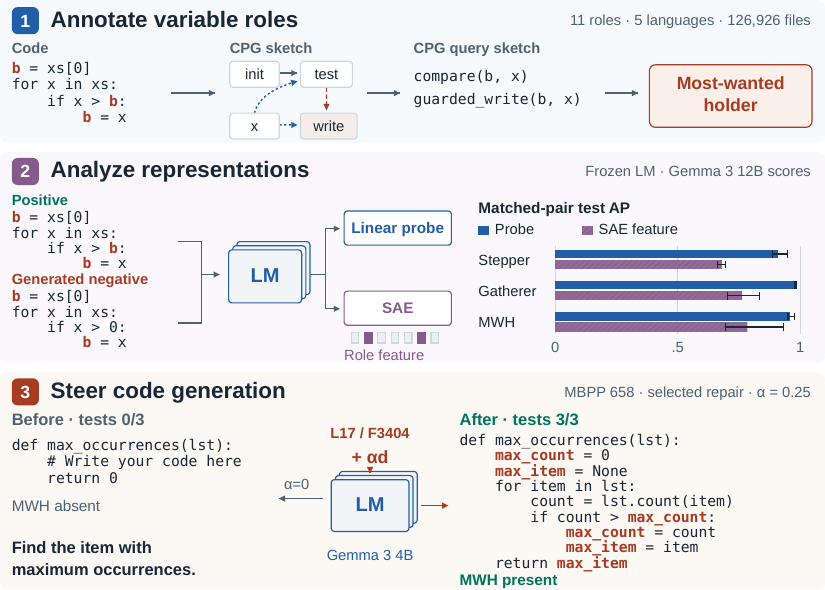}
    \caption{
        Variable roles connect code annotation, representation analysis, and steering.
        \textbf{(1)} Code Property Graph (CPG) queries annotate all eleven Sajaniemi variable roles at scale.
        \textbf{(2)} Probes and SAE features recover role information; bars show matched-pair test average precision (AP) for Gemma~3~12B (four-layer mean and range).
        Code and graph sketches illustrate the method.
        \textbf{(3)} In Gemma~3~4B, steering a most-wanted-holder (MWH) feature changes a stub into a working search.
    }
    \label{fig:overview}
\end{figure}

We combine program analysis with sparse autoencoder (SAE) features and probes, contributing:
\begin{itemize}
\item \textbf{Large-scale multilingual dataset and benchmark.}
We introduce a dataset of 126,926 natural-code files across five programming languages, to our knowledge the first to automatically annotate all eleven Sajaniemi variable roles at this scale.
We construct a benchmark with matched and mined negatives, LLM-reviewed role-changing edits, and role-preserving target renaming to measure recoverability under controlled comparisons.
We~will publicly release the benchmark, annotations, evaluation splits, and extraction pipeline, with instructions for reconstructing the corpus.
\item \textbf{Recoverability and SAE fidelity.}
Across five model--SAE systems, we compare residual probes, reconstruction probes, individual features, and probes combining 64 features at observation points after the selected operation.
Controlled edits test sensitivity beyond matched prediction; language, implementation, and readout comparisons qualify recoverability.
\item \textbf{Limits of intervention-based control.}
We measure role induction and suppression alongside functional correctness and random-direction controls.
Annotation changes are substantially more frequent than changes accompanied by passing tests; Figure~\ref{fig:overview} illustrates a selected successful repair.
\end{itemize}

\begin{figure}[t]
\centering
\setlength{\abovecaptionskip}{6pt}
\includegraphics[width=\columnwidth]{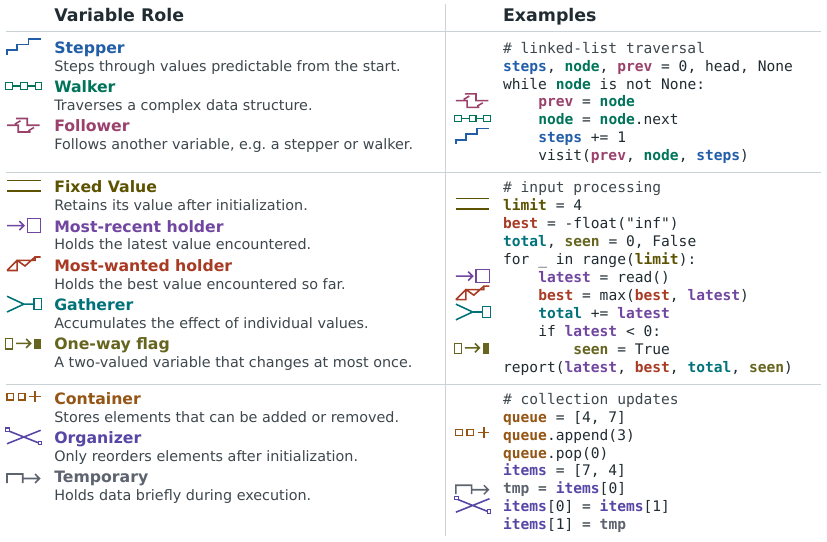}
\caption{Overview of conceptual variable roles according to
\citet{sajaniemi2002visualizing} and \citet{sajaniemi2005experts}.
Matching colors and symbols link definitions to variable occurrences in Python
code. Examples are deliberately simple, chosen for illustration.
Appendix~\ref{app:real-world-roles} shows more complex examples from real-world
code.}
\label{fig:taxonomy}
\end{figure}

\section{Background: Variable Roles in Program Comprehension}
\label{sec:background}

\paragraph{Program comprehension and programming knowledge.}
Research on human program comprehension distinguishes representations of operations and control flow from a \emph{situation model} of program goals, with task-dependent prominence \citep{pennington1987representations}.
\emph{Programming plans} describe familiar patterns of operations that accomplish particular goals \citep{soloway1984knowledge}.
Sajaniemi's roles bring this perspective to individual variables through \emph{variable plans}, or stereotyped uses of variables.
The taxonomy originated in textbook programs; subsequent card sorting and interviews with professional programmers recovered groupings consistent with the roles, alongside variation in behavioral interpretations \citep{sajaniemi2005experts}.

\paragraph{Variable roles as behavioral concepts.}
A role characterizes a variable through its successive values and how they are obtained \citep{sajaniemi2002visualizing,sajaniemi2005experts}: a \emph{stepper} follows a predictable progression, a \emph{gatherer} combines contributions, and a \emph{most-wanted holder} retains the best candidate encountered.
Figure~\ref{fig:taxonomy} summarizes all eleven Sajaniemi variable roles; Appendix~\ref{app:role-realizations} illustrates alternative implementations.
Roles describe variable behavior, so role preservation differs from program equivalence: minimum and maximum computations can share a role, whereas equivalent programs can use variables with different roles.
Although defined independently of the models, these roles can correlate with identifiers or syntax; testing their representation therefore requires controls for such surface cues.

\paragraph{Variable roles in neural models of code.}
\citet{hussain2023roles} detect steppers and walkers and inject their roles into Code2Seq representations.
\citet{demirtas2025planning} use LLMs to detect programming plans in student programs.
These studies use programming abstractions as input features or prediction targets; we compare role recoverability under controlled source edits with intervention outcomes during generation.
This comparison requires annotations defined independently of model activations and linked to individual variables in natural code.

\section{Annotating Variable Roles at Scale}

To evaluate all eleven roles in multilingual natural code, we develop an automatic annotation procedure with explicit behavioral contracts.

\paragraph{Automatic role extraction.}
Scaling annotation requires operational definitions of these cognitive abstractions.
Prior systems cover subsets of the taxonomy, use execution traces, or adopt alternative definitions (Table~\ref{tab:variable_role_extraction}).
We operationalize all eleven Sajaniemi variable roles as explicit \emph{role contracts} over \emph{Joern} Code Property Graphs (CPGs), which represent syntax, control flow, and data dependencies \citep{yamaguchi2014cpg}.
Queries recover initializations, updates, and uses, whose relationships are checked against the contracts.
For example, the \emph{most-wanted holder} contract links a candidate comparison to conditional replacement of the current value, with consistency checks on initialization and other writes.
These static checks require neither execution nor a buildable project, enabling annotation of isolated source files.
Appendix~\ref{app:role-contracts} gives the contracts and pseudocode.

\begin{table}[t]
\centering
\caption{Automatic variable-role analysis and reported corpus sizes.
All but Demyanova et al. use Sajaniemi's taxonomy.
NR: not reported; vars.: role-labeled variables; NN: neural classifier.
${}^{\dagger}$: role-definition sample.
Appendix~\ref{app:prior-dataset-sizes} details annotation units, breakdowns,
and comparability.}
\label{tab:variable_role_extraction}
\begingroup
\footnotesize
\setlength{\tabcolsep}{2pt}
\renewcommand{\arraystretch}{1.05}
\begin{tabular*}{\linewidth}{@{\extracolsep{\fill}}lllcrr@{}}
\toprule
\textbf{Work} & \textbf{Signal} & \textbf{Languages} & \textbf{Roles} &
\textbf{Corpus} & \textbf{Labeled data} \\
\midrule
\citet{gerdt2004automatic} & Static analysis & Pascal & 3 & NR & NR \\
\addlinespace[2pt]
\citet{bishop2005assessing} & Static analysis & Java & Checking & 17 programs & NR \\
\addlinespace[2pt]
\citet{gerdt2006web} & Static + learning & Pascal & 9 & 87 programs & 305 vars. \\
\addlinespace[2pt]
\citet{demyanova2013variableRoles} & Static analysis & C & 14 custom & 5.2 KLOC${}^{\dagger}$ & NR \\
\addlinespace[2pt]
\citet{lui2020recognition} & Traces + NN & C & 4 classes & 110 programs & 8,907 cases \\
\addlinespace[2pt]
\citet{hussain2023roles} & Syntax rules & Java & 2 & 14,022,318 methods & 1,438,035 vars. \\
\midrule
\textbf{Ours} & \textbf{Static CPG} & \textbf{C, C++, JS,} & \textbf{11} & \textbf{126,926 files} & \textbf{1,737,758 vars.} \\
 & \textbf{analysis} & \textbf{Python, Ruby} & & & \\
\bottomrule
\end{tabular*}
\endgroup
\end{table}

\paragraph{Constructing a multilingual natural-code dataset.}
We apply these contracts to CodeParrot GitHub-Code, combining an expanded annotation run with an earlier role-enriched subset, including Ruby.
Deduplicating sources and matching declarations yields 126,926 files across five languages and 3,525,530 variables, of which 1,737,758 have at least one extracted role.
Variables lacking a target annotation serve as candidate negatives, including any role instances missed by the extractor.
For evaluation, we form five approximately equal folds stratified by role and language, keeping exact-source and detected clone groups intact.
Appendix~\ref{app:dataset} reports role and language distributions.

\paragraph{Annotation validity and evaluation scope.}
Source inspection and implementation checks informed contract refinement, but coverage remains limited to supported static patterns. A human evaluation on 660 stratified variables (Appendix~\ref{app:variable_extractor_human_eval}) assesses annotation quality, although its sampling design does not support corpus-wide precision or recall estimates.
Counterfactual evaluation additionally uses case-level LLM review of original behavior and proposed edits (Appendix~\ref{app:evaluation-protocol}). We therefore distinguish annotation accuracy on the evaluated subset from automatic-label prediction and sensitivity to reviewed edits.

\section{Do Language Models Represent Variable Roles?}
\label{sec:decodability}

We first measure recoverability of role annotations, then test sensitivity to role-changing edits and stability under role-preserving renaming.

\begin{figure}[t]
    \centering
    \setlength{\abovecaptionskip}{6pt}
    \includegraphics[width=\linewidth]{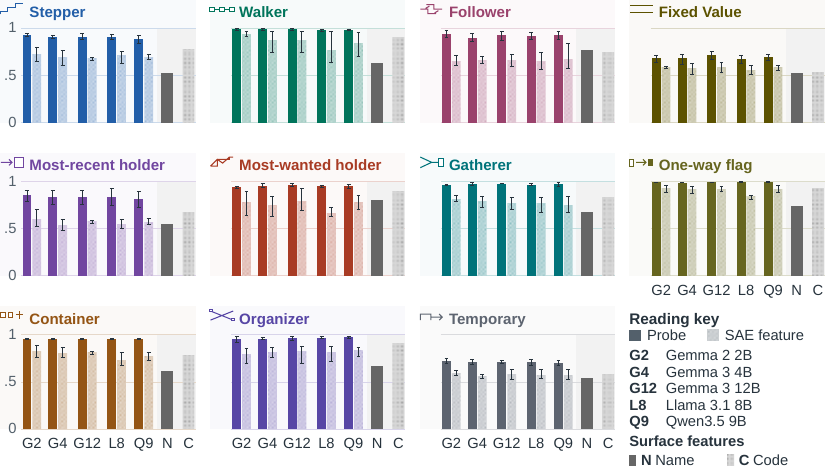}
    \caption{
        \textbf{Model representations support role recovery beyond the tested surface baselines.}
        Residual probes attain higher mean test average precision (AP) than both surface baselines and individual SAE features for every role and model.
        Evaluation uses matched C, C++, JavaScript and Python pairs from one source-group split.
        Model bars average AP over four prescribed layers; whiskers show the layer range.
        Gray baselines use character $n$-grams of names (N) or code prefixes through the causal readout (C).
        Role colors and symbols follow Figure~\ref{fig:taxonomy}; solid and hatched bars denote Probe and validation-selected SAE feature.
    }
    \label{fig:role_ap}
\end{figure}

\paragraph{Matched prediction and diagnostic readout.}
To reduce surface differences, we pair each annotated positive with a variable lacking that annotation, matching language and declaration kind and favoring similar surface characteristics.
The test set contains 4,238 balanced pairs across eleven roles in C, C++, JavaScript, and Python.
Because assignment targets often precede their determining expressions, we read the state of a target reference inserted after the selected statement or loop/conditional header.
Appendix~\ref{app:evaluation-protocol} records selection, exclusions, and fitting; Appendix~\ref{app:negative-strategies} illustrates the contrasts.

\paragraph{Four predictors separate feature restriction from reconstruction.}
We evaluate Gemma~2~2B, Gemma~3~4B/12B, Llama~3.1~8B, and Qwen3.5~9B at four layers each.
SAEs encode residual activations as sparse feature vectors and reconstruct them through a linear decoder \citep{huben2024sparse,gao2024scaling}.
Ridge probes on residual activations, SAE reconstructions, and 64 SAE features separate reconstruction fidelity from restricted feature readout; a single-feature predictor tests individual coordinates.
Training fixes standardization, feature ranking, and polarity; validation selects regularization, the single feature, and decision thresholds.
The \emph{name} and \emph{code} surface baselines are TF--IDF logistic classifiers over character $n$-grams of the target name and the code prefix through the diagnostic token, respectively.
Fits are frozen before evaluation; Appendix~\ref{app:sae-details} lists SAE releases.

\paragraph{Matched recoverability exceeds the tested lexical baselines.}
Residual probes exceed both surface baselines and individual SAE features in every role--model comparison (Figure~\ref{fig:role_ap}).
Averaging roles, models, and layers equally, average precision (AP) is $0.90$ for residual probes and $0.72$ for single features; constant-score AP is $0.50$.
Code features are competitive with individual SAE features: for most-wanted holder, their AP is $0.90$, versus $0.75$ for single features and $0.95$ for residual probes.
Reconstruction and top-64 probes close much of the single-feature gap (Table~\ref{tab:representation-controls}), showing the cost of restricting the readout to one coordinate.

\begin{table}[t]
\centering
\small
\setlength{\tabcolsep}{5pt}
\begin{tabular}{lcc}
\toprule
Predictor & Matched AUROC & Counterfactual paired accuracy \\
\midrule
Residual probe & 0.906--0.913 & 0.813--0.877 \\
SAE reconstruction probe & 0.857--0.881 & 0.739--0.784 \\
Top-64 SAE probe & 0.853--0.878 & 0.722--0.753 \\
Single SAE feature & 0.722--0.782 & 0.622--0.713 \\
\bottomrule
\end{tabular}
\caption{\textbf{Multiple-feature readouts close much of the single-feature gap; reconstruction probes have lower counterfactual sensitivity.}
Values span five model means, each weighting eleven roles and four prescribed layers equally, with matched-negative training.
Ranges show variation across systems; paired uncertainty for reconstruction-minus-residual differences is reported in the text.
Counterfactual accuracy compares each original only with its corresponding source-reviewed edit.
}
\label{tab:representation-controls}
\end{table}

\begin{figure}[t]
\centering
\includegraphics[width=\linewidth]{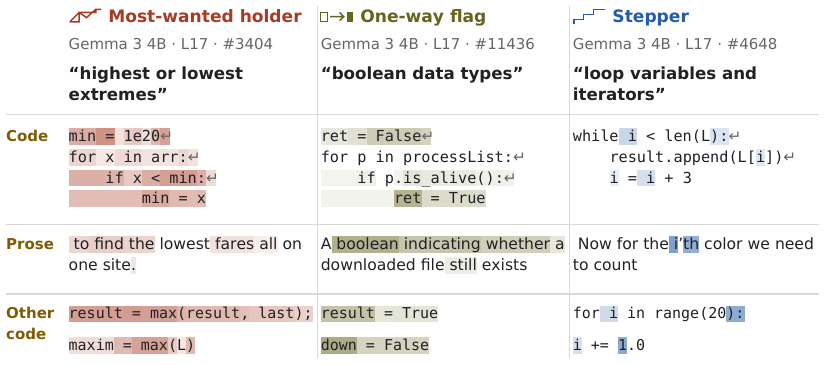}
\caption{\textbf{Role-associated latents connect code updates to broader extrema, Boolean, and indexing concepts.}
Role colors and symbols follow Figure~\ref{fig:taxonomy}; descriptions and prose come from Neuronpedia, and code from our training corpus.
Highlight intensity indicates latent activation strength on a linear scale within each latent; return marks denote newlines.
Code activations use standalone snippets; prose windows retain their original Neuronpedia contexts.}
\label{fig:role-latent-pair}
\end{figure}

\paragraph{Counterfactual ranking and classification give different answers.}
To test whether matched recoverability tracks defining behavior, we evaluate edits that remove the target role.
On 1,698 LLM-reviewed pairs from 948 source groups, residual probes rank originals above edited negatives with paired accuracy $0.813$--$0.877$; ties receive half credit.
The code-prefix baseline reaches $0.750$, while unchanged names give the name baseline $0.500$.
Reconstruction probes fall 4.0--12.3 percentage points below residual probes, with all five paired 95\% source-group bootstrap intervals excluding zero: reconstruction retains less counterfactual sensitivity.
At~the validation thresholds, however, residual probes classify 68--71\% of edited negatives as positive, yielding balanced accuracy of 0.610--0.637.
The scores therefore respond to role-changing edits more reliably than their binary decisions reject the edited examples.

\paragraph{Neutral target renaming preserves most ranking performance.}
As a complementary role-preserving control, renaming both members of 100 test pairs per role reduces residual AUROC (area under the ROC curve) by 0.6--1.7 percentage points, to 0.878--0.887.
On these same pairs, code-surface AUROC changes from 0.740 to 0.723 and name-surface AUROC from 0.602 to 0.492.
Renaming preserves other identifiers, comments, and implementation cues, isolating sensitivity to the target spelling.
Appendix~\ref{app:additional-role-analyses} reports uncertainty and decision changes; Appendix~\ref{app:behavioral-experiments} evaluates prompted recognition.

\paragraph{Readout and implementation condition the signal.}
The diagnostic readout also matters because it follows the selected operation.
At the second prescribed layer, residual counterfactual paired accuracy is 0.495--0.594 at the native identifier, 0.790--0.915 at statement end, and 0.856--0.920 at the inserted identifier.
Separately fitted readouts thus expose much of this sensitivity at later unmodified positions.
Even at the diagnostic readout, recoverability varies: with multilingual training, stepper AUROC ranges from 0.682 in JavaScript to 0.972 in Python; implementation contrasts can also be confounded by language (Appendix~\ref{app:additional-role-analyses}).

Figure~\ref{fig:role-latent-pair} illustrates the broader extrema, Boolean, and indexing associations of selected features.
These contexts aid interpretation but do not establish role specificity (Appendix~\ref{app:feature-contexts}).
Neither contextual associations nor counterfactual sensitivity establish control; we next test interventions during generation.

\section{Do Variable Role Representations Influence Code Generation?}
\label{sec:causal-influence}

\suppressfloats[t]
\begin{figure}[t]
\centering
\includegraphics[width=\linewidth]{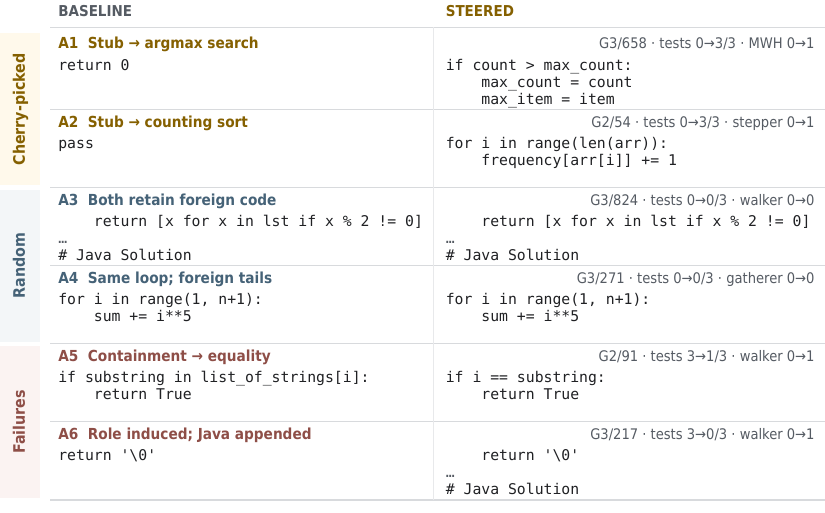}
\caption{\textbf{Positive steering can repair stubs but can also break correct programs.}
Each pair shows held-out MBPP excerpts; test counts are passing supplied assertions out of three, and role bits denote extractor output.
G2/G3 denote Gemma~2~2B/Gemma~3~4B; MWH denotes most-wanted holder.
The selected repairs and failures are illustrative, while the random cases come from a uniform draw without replacement.
Outcomes evaluate complete outputs, including any omitted foreign-language tails.}
\label{fig:steering-examples}
\end{figure}

To test control beyond recoverability, we apply activation addition and direction removal \citep{turner2023activation,arditi2024refusal,ferrando2024entity,tahimic2025codecorrectness} on Mostly Basic Programming Problems (MBPP) \citep{austin2021program}.
The SAE-feature experiments cover Gemma~2~2B and Gemma~3~4B/12B; a separate comparison normalizes probe and SAE directions under a correctness constraint.
For each role $r$, binary extractor labels on reference ($R_{\mathrm{ref}}$) and unsteered ($R_0$) solutions define cohorts.
These labels describe implementations: a correct alternative may use a different role.
Target and random directions share eligible tasks; Pass@1 measures the fraction passing the supplied tests.

\begin{figure}[t]
\centering
\includegraphics[width=\columnwidth]{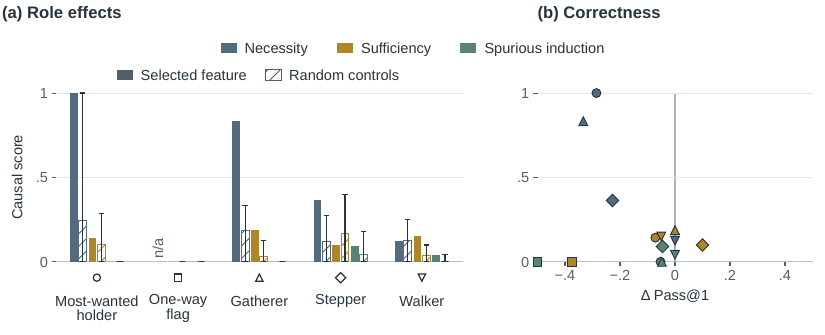}
\caption{
\textbf{Causal effects of representative role-associated SAE features in Gemma 2 2B.}
(a) Scores for one selected feature per role and same-layer random SAE controls (hatched means; range whiskers); n/a denotes undefined necessity.
(b) Causal scores versus changes in MBPP Pass@1; shapes match role labels in (a).
Removal suggests causal involvement with limited specificity and substantial correctness costs; positive steering provides weak evidence of sufficiency.
}
\label{fig:causal_scores_and_pass_at_1}
\end{figure}

\paragraph{Reference-conditioned induction.}
For feature $f$, we add $\alpha d_f$, where $d_f=W_{\mathrm{dec}}[f,:]$, to the residual stream at every generated token; $R_\alpha$ denotes the resulting role annotation.
The operational score \emph{sufficiency} measures induction of an annotation present in the reference but absent from the baseline:
\[
    \mathrm{Sufficiency}_r
    =
    P(R_{\alpha^\star}=1
    \mid R_{\mathrm{ref}}=1,R_0=0).
\]
Eligible examples are split equally into calibration and test sets; $\alpha^\star$ maximizes calibration sufficiency over a predefined strength grid and is fixed for testing.
Reference-negative induction, denoted \emph{spurious induction} in Figure~\ref{fig:causal_scores_and_pass_at_1}, uses a separate cohort:
\[
    \mathrm{SpuriousInduction}_r
    =
    P(R_{\alpha^\star}=1
    \mid R_{\mathrm{ref}}=0,R_0=0).
\]
Because reference-negative induction may yield a valid alternative, both cohorts require separate correctness evaluation.

\paragraph{Reference-conditioned suppression.}
To test suppression of roles present in both reference and baseline, we normalize the decoder direction to $\hat d_f$ and orthogonalize matrices writing to the residual stream \citep{arditi2024refusal,ferrando2024entity,tahimic2025codecorrectness}:
\[
    W_{\mathrm{out}}'
    =
    W_{\mathrm{out}}
    -
    (W_{\mathrm{out}}\hat d_f^\top)\hat d_f.
\]
The operational score \emph{necessity} is the frequency of role absence after removal, denoted by $R_{\mathrm{ort}}=0$:
\[
    \mathrm{Necessity}_r
    =
    P(R_{\mathrm{ort}}=0
    \mid R_{\mathrm{ref}}=1,R_0=1).
\]
With no fitted strength, removal uses all eligible examples.
Unlike generation-time addition, global removal also changes prompt processing; effect sizes reflect different perturbation scopes.

\suppressfloats[t]
\begin{table*}[t]
\centering
\small
\setlength{\tabcolsep}{5pt}
\begin{tabular}{llrrrr}
\toprule
& & \multicolumn{2}{c}{Gemma 2 2B} & \multicolumn{2}{c}{Gemma 3 4B} \\
\cmidrule(lr){3-4}\cmidrule(lr){5-6}
Role & Representative syntax & Pos. steer. & Orth. & Pos. steer. & Orth. \\
\midrule
Gatherer & Augmented assignment & $+1.9$ (107) & $-58.8$ (17) & $+2.3$ (44) & $-50.0$ (4) \\
Most-wanted holder & \texttt{if} statement & $-4.3$ (92) & $-35.0$ (20) & $+17.6$ (17) & $-75.0$ (4) \\
One-way flag & Boolean literal & $-5.9$ (17) & -- & $-11.1$ (18) & -- \\
Stepper & \texttt{range} call & $+20.3$ (59) & $-20.0$ (40) & $+9.3$ (54) & $-63.6$ (11) \\
Walker & Subscript expression & $-2.5$ (40) & $+0.0$ (8) & $-5.1$ (39) & $+0.0$ (10) \\
\bottomrule
\end{tabular}
\caption{\textbf{Role-adjacent syntax changes after intervention.}
Cells give intervention-minus-baseline changes (percentage points) in program frequencies of Tree-sitter patterns, with counts of pairs with two clean parses in parentheses.
Patterns are structural correlates, not role definitions or evidence of full-role sufficiency.
Cohorts and baselines differ by intervention; dashes mark unavailable removal generations.
Counts repeat tasks across features; Appendix~\ref{app:causal-consolidated} reports parse survival.}
\label{tab:causal-role-syntax}
\end{table*}

\paragraph{Removal suggests causal involvement with limited specificity.}
Some Gemma~2~2B features have high necessity (Figure~\ref{fig:causal_scores_and_pass_at_1}).
Gatherer suppression exceeds its random-control range, whereas most-wanted holder's score of $1.00$ ties its control maximum.
Their Pass@1 losses ($-0.33$ and $-0.29$) suggest suppression may partly reflect broader generation disruption, limiting evidence of role-specific control.

This correctness cost persists across all selected SAE features: removal suppresses annotations in 41/91, 72/109, and 131/187 target--task outcomes for Gemma~2~2B, Gemma~3~4B, and Gemma~3~12B.
Only 2/91, 4/109, and 15/187 outcomes also pass the supplied tests; requiring both baseline and intervention to pass leaves 0, 3, and 13.
Rates of annotation changes with passing tests exceed matched random controls by 1.3--3.7 percentage points for removal and 0.9--1.9 for addition, but all six task-grouped bootstrap 95\% intervals include zero (Appendix~\ref{app:causal-consolidated}).
These joint outcomes show no clear advantage over random controls.

\paragraph{Weak sufficiency, partial structural effects.}
Positive steering yields nonzero sufficiency scores of only $0.10$--$0.19$, even where they exceed random-control ranges (Figure~\ref{fig:causal_scores_and_pass_at_1}); low reference-negative induction likewise reflects weak effects.
Yet Table~\ref{tab:causal-role-syntax} shows increased stepper-associated \texttt{range} calls under positive steering in both models, with smaller increases in gatherer-associated augmented assignments.
These partial structural effects do not establish full-role sufficiency; other shifts are mixed.
Only pairs with two clean parses contribute; Gemma~3~4B removal retains just 4/30 gatherer and 11/48 stepper pairs, from two and five tasks, limiting generalization.

\paragraph{Correctness-constrained calibration usually selects zero.}
To test whether calibration can avoid these correctness costs, we constrain correctness in a normalized probe/SAE comparison; ten of eleven completed model--role cells select no intervention (Appendix~\ref{app:matched-steering}).
In the remaining cell, each target direction produces one induction accompanied by passing tests among five eligible tasks, matching the random residual direction.
Figure~\ref{fig:steering-examples} illustrates repairs and failures, with selected successes and unfiltered random cases documented in Appendix~\ref{app:steering-examples}.

\section{Related work}
\label{sec:related}

These distinctions connect work on behavior, representations, and interventions (Appendix~\ref{app:related-work}).

\paragraph{Behavioral robustness and semantic priors.}
Semantics-preserving transformations test reliance on surface cues.
ReCode finds reduced generation robustness after changes to identifiers, syntax, formatting, and docstrings \citep{wang2023recode}; related mutations destabilize program-output predictions \citep{orvalho2025mutations}.
Masked-code models retain predictive ability after operand and branch reordering and identifier renaming, suggesting semantic learning \citep{ahmed2023understanding}.

\paragraph{Representations of programming concepts.}
Probes recover syntax and structure \citep{karmakar2021probing,wan2022structural}, including complete abstract syntax trees \citep{hernandezlopez2022astprobe}, and intermediate execution states in synthetic-language models \citep{jin2024semantics}.
Whereas probes target predefined properties, CoCoA discovers token-representation clusters and interprets their lexical, syntactic, and semantic content through static analysis and LLM annotation \citep{sharma2025latent}.
Variable roles let us compare residual and SAE recoverability under source edits, separating association from sensitivity to defining behavior.

\paragraph{Shared representations across programming and natural languages.}
Cross-language studies test dependence on language-specific syntax.
The same attention heads convey variable-type information across three programming languages \citep{north2025beyond}, while language-specific and language-agnostic neurons coexist within a model \citep{yin2026neuronguided}.
Intermediate-layer decoding and neuron ablations also reveal overlap with English \citep{kargaran2025shared}.
Our multilingual fits complement transfer studies by measuring variation in role recovery within each evaluated language.

\paragraph{Causal and mechanistic analyses.}
Coding failures can coexist with useful computations: steering restores disrupted type predictions \citep{lucchetti2025types}, while synthetic parentheses analyses trace errors to interference between reliable and unreliable components \citep{rai2025interference}.
Yet SAE features predictive of code correctness need not support effective steering \citep{tahimic2025codecorrectness}.
We extend this distinction to role annotations, comparing changes, joint passing outcomes, and matched random controls.

\section{Conclusion, Limitations, and Future Work}
\label{sec:conclusion}
Residual probes recover operational role annotations beyond tested lexical baselines.
Combining SAE features narrows the single-feature gap, but reconstruction probes retain less sensitivity to reviewed role-changing edits.
Removal suggests causal involvement with limited specificity and correctness costs, while positive steering yields weak full-role induction despite structural shifts.

\paragraph{Limitations.}
Human evaluation shows that annotation quality varies substantially across roles, particularly for most-recent holder (Appendix~\ref{app:variable_extractor_human_eval}). These estimates come from a stratified sample and do not establish corpus-wide precision or recall; inter-annotator reliability also remains unmeasured.
Decoding covers four layers per model and one source-group split, with uncertainty conditional on fitted predictors.
The corpus includes development data without repository-level holdout, and target renaming retains other lexical cues.
Interventions use small overlapping cohorts and finite prompt-visible tests, with unresolved historical scale provenance (Appendix~\ref{app:causal-consolidated}).

\paragraph{Future work.}
Our corpus combines natural-code complexity with independently defined roles and controlled source edits, offering a semi-controlled setting for studying conceptual abstraction beyond toy tasks.
Future work can test whether role representations transfer to unseen languages and implementations, combine into higher-level programming plans, and support interventions that change variable behavior while preserving correctness.

\section{Use of AI Assistance}
\label{sec:ai-use}

We used large language models and coding agents to assist literature searches, manuscript drafting and revision, visualization code, and experimental protocols.
The authors reviewed all references, procedures, figures, and manuscript content before inclusion.
LLM reviewers also assessed counterfactual source edits, recording original-role validity, edited behavior, and rejection reasons; Appendix~\ref{app:evaluation-protocol} distinguishes this procedure from human annotation validation.

AI assistance was more extensive in implementing the variable-role extractor, whose multilingual static-analysis rules combine syntax, control flow, and data flow.
The authors specified behavioral requirements and rejection conditions for each role (Appendix~\ref{app:role-contracts}); coding agents implemented these contracts using Joern CPG facts and the existing extraction infrastructure.
We inspected implementations and annotations manually on source-code examples and automatically through checks on annotations and intermediate variable facts.
Failures and false positives or negatives informed contract refinements, followed by implementation updates and re-evaluation until both stabilized.
The authors controlled the choice of taxonomy, operational definitions, final annotation rules, experimental design, interpretation, and manuscript.

\bibliographystyle{iclr2027_conference}
\bibliography{references}

\clearpage
\appendix
\etocdepthtag.toc{appendix}
\begingroup
\hypersetup{hidelinks,linktoc=all}
\etocsettagdepth{main}{none}
\etocsettagdepth{appendix}{subsection}
\etocstandardlines
\etocsettocstyle{\section*{Appendix contents}}{}
\pdfbookmark[0]{Appendix contents}{appendix-contents}
\tableofcontents
\endgroup
\clearpage

\section{Variable roles in real-world code}
\label{app:real-world-roles}

The simple examples in Figure~\ref{fig:taxonomy} isolate each role's behavior.
The six excerpts below cover all eleven Sajaniemi variable roles in natural code from our annotated corpus.
Common leading indentation is removed; long lines wrap for display.

\begingroup
\raggedbottom
\definecolor{rwfixed}{HTML}{5C5200}
\definecolor{rwstepper}{HTML}{225EA8}
\definecolor{rwwalker}{HTML}{00745B}
\definecolor{rwfollower}{HTML}{9A426C}
\definecolor{rwrecent}{HTML}{7146A0}
\definecolor{rwwanted}{HTML}{A83B24}
\definecolor{rwgatherer}{HTML}{00737A}
\definecolor{rwflag}{HTML}{66651F}
\definecolor{rwcontainer}{HTML}{945515}
\definecolor{rworganizer}{HTML}{5746A5}
\definecolor{rwtemporary}{HTML}{5D6470}
\definecolor{rwcomment}{HTML}{596168}
\definecolor{rwrule}{HTML}{CBD3D7}
\lstdefinestyle{realworldroles}{language=Python,
  basicstyle=\fontsize{8.2}{9.5}\selectfont\ttfamily,
  keywordstyle=\bfseries,commentstyle=\color{rwcomment},
  stringstyle=\color{rwcomment},showstringspaces=false,
  columns=fullflexible,keepspaces=true,alsoletter={_},tabsize=4,
  numbers=left,numberstyle=\tiny\color{rwcomment},numbersep=7pt,
  xleftmargin=15pt,frame=tb,framerule=0.3pt,rulecolor=\color{rwrule},
  framesep=4pt,aboveskip=5pt,belowskip=5pt,
  breaklines=true,breakatwhitespace=false,breakindent=12pt,
  postbreak=\mbox{\normalfont\scriptsize\textcolor{rwcomment}{$\hookrightarrow$}\space}}

\par\medskip\noindent
\begin{minipage}{\linewidth}
\textbf{Streaming an application release.}\hfill
{\small\href{https://github.com/nextcloud/appstore/blob/744859a5e40c7e0f49df98f8aeab1c451815ad01/nextcloudappstore/api/v1/release/downloader.py}{Nextcloud App Store}}
\begin{lstlisting}[style=realworldroles,
  emph={max_size},emphstyle=\bfseries\color{rwfixed},
  emph={[2]size},emphstyle={[2]\bfseries\color{rwgatherer}},
  emph={[3]finished},emphstyle={[3]\bfseries\color{rwflag}},
  emph={[4]msg},emphstyle={[4]\bfseries\color{rwtemporary}}]
def _stream_to_file(self, file: Any, max_size: int,
                    req: requests.Response) -> None:
    # start streaming download
    finished = False
    try:
        size = 0
        for chunk in req.iter_content(1024):
            file.write(chunk)
            size += len(chunk)
            if size > max_size:
                msg = 'Downloaded archive is bigger than the ' \
                      'allowed %i bytes' % max_size
                raise MaximumDownloadSizeExceededException(msg)
        finished = True
    finally:
        # in case any errors occurred, get rid of the file
        file.close()
        if not finished:
            os.remove(file.name)
\end{lstlisting}
\small
The \emph{fixed value} \textcolor{rwfixed}{\texttt{max\_size}} bounds the download.
The \emph{gatherer} \textcolor{rwgatherer}{\texttt{size}} adds input-dependent chunk lengths, and the \emph{one-way flag} \textcolor{rwflag}{\texttt{finished}} changes to true only after streaming completes.
The \emph{temporary} \textcolor{rwtemporary}{\texttt{msg}} holds an exception message.
The \texttt{finally} block uses the flag to clean up after an error.
\end{minipage}

\par\bigskip\noindent
\begin{minipage}{\linewidth}
\textbf{Applying byte-level delta operations.}\hfill
{\small\href{https://github.com/jelmer/subvertpy/blob/c1b317d2a65ce63720cf3edd488e868532918add/subvertpy/delta.py}{Subvertpy}}
\begin{lstlisting}[style=realworldroles,
  emph={action},emphstyle=\bfseries\color{rwwalker},
  emph={[2]i},emphstyle={[2]\bfseries\color{rwstepper}},
  emph={[3]tview},emphstyle={[3]\bfseries\color{rwcontainer}}]
def txdelta_apply_ops(src_ops, ops, new_data, sview):
    """Apply txdelta operations to a source view.

    :param src_ops: Source operations, ignored.
    :param ops: List of operations (action, offset, length).
    :param new_data: Buffer to fetch fragments with new data from
    :param sview: Source data
    :return: Result data
    """
    tview = bytearray()
    for (action, offset, length) in ops:
        if action == TXDELTA_SOURCE:
            # Copy from source area.
            tview.extend(sview[offset:offset+length])
        elif action == TXDELTA_TARGET:
            for i in range(length):
                tview.append(tview[offset+i])
        elif action == TXDELTA_NEW:
            tview.extend(new_data[offset:offset+length])
        else:
            raise Exception("Invalid delta instruction code")
    return tview
\end{lstlisting}
\small
The \emph{walker} \textcolor{rwwalker}{\texttt{action}} takes the operation code from each tuple in \texttt{ops}.
The \emph{stepper} \textcolor{rwstepper}{\texttt{i}} advances through the range of each target-copy operation.
The \emph{container} \textcolor{rwcontainer}{\texttt{tview}} grows through bulk extension and single-byte appends.
These roles coexist in a dispatch loop for different copy operations.
\end{minipage}

\par\bigskip\noindent
\begin{minipage}{\linewidth}
\textbf{Reading blocks and emitting delta windows.}\hfill
{\small\href{https://github.com/jelmer/subvertpy/blob/c1b317d2a65ce63720cf3edd488e868532918add/subvertpy/delta.py}{Subvertpy}}
\begin{lstlisting}[style=realworldroles,
  emph={text},emphstyle=\bfseries\color{rwrecent}]
def send_stream(stream, handler, block_size=DELTA_WINDOW_SIZE):
    """Send txdelta windows that create stream to handler

    :param stream: file-like object to read the file from
    :param handler: txdelta window handler function
    :return: MD5 hash over the stream
    """
    hash = md5()
    text = stream.read(block_size)
    if not isinstance(text, bytes):
        raise TypeError("The stream should read out bytes")
    while text:
        hash.update(text)
        window = (0, 0, len(text), 0, [(TXDELTA_NEW, 0, len(text))], text)
        handler(window)
        text = stream.read(block_size)
    handler(None)
    return hash.digest()
\end{lstlisting}
\small
The \emph{most-recent holder} \textcolor{rwrecent}{\texttt{text}} is replaced by each block read from the stream.
Each block updates the digest and enters a delta window before the next read.
A type check and a final callback surround this acquisition--use cycle.
\end{minipage}

\par\bigskip\noindent
\begin{minipage}{\linewidth}
\textbf{Monitoring peak memory usage.}\hfill
{\small\href{https://github.com/ianozsvald/ipython_memory_usage/blob/443ee344e6e6d808a8315b35e8bbbedffaf5ad3f/ipython_memory_usage.py}{ipython\_memory\_usage}}
\begin{lstlisting}[style=realworldroles,
  emph={peak_memory_usage},emphstyle=\bfseries\color{rwwanted}]
def during_execution_memory_sampler():
    import time
    import memory_profiler
    global keep_watching, peak_memory_usage
    peak_memory_usage = -1
    keep_watching = True

    n = 0
    WAIT_BETWEEN_SAMPLES_SECS = 0.001
    MAX_ITERATIONS = 60.0 / WAIT_BETWEEN_SAMPLES_SECS
    while True:
        mem_usage = memory_profiler.memory_usage()[0]
        peak_memory_usage = max(mem_usage, peak_memory_usage)
        time.sleep(WAIT_BETWEEN_SAMPLES_SECS)
        if not keep_watching or n > MAX_ITERATIONS:
            # exit if we've been told our command has finished or if it has run
            # for more than a sane amount of time (e.g. maybe something crashed
            # and we don't want this to carry on running)
            if n > MAX_ITERATIONS:
                print("{} SOMETHING WEIRD HAPPENED AND THIS RAN FOR TOO LONG, THIS THREAD IS KILLING ITSELF".format(__file__))
            break
        n += 1
\end{lstlisting}
\small
The \emph{most-wanted holder} \textcolor{rwwanted}{\texttt{peak\_memory\_usage}} retains the largest observed usage during each sampling run.
Its monitoring loop also handles a stop signal, a sampling delay, and a timeout.
\end{minipage}

\par\bigskip\noindent
\begin{minipage}{\linewidth}
\textbf{Searching parent directories for configuration.}\hfill
{\small\href{https://github.com/modesttree/Projeny/blob/4c11ab8f3d2c91765bdf50766f39b2df1d208467/Source/prj/main/OpenInVisualStudio.py}{Projeny}}
\begin{lstlisting}[style=realworldroles,
  emph={lastParentDir},emphstyle=\bfseries\color{rwfollower}]
def findConfigPath(filePath):
    lastParentDir = None
    parentDir = os.path.dirname(filePath)

    while parentDir and parentDir != lastParentDir:
        configPath = os.path.join(parentDir, Prj.ConfigFileName)

        if os.path.isfile(configPath):
            return configPath

        lastParentDir = parentDir
        parentDir = os.path.dirname(parentDir)

    assertThat(False, "Could not find Prj config path starting at path {0}", filePath)
\end{lstlisting}
\small
The \emph{follower} \textcolor{rwfollower}{\texttt{lastParentDir}} preserves \texttt{parentDir}'s old value before the next parent-directory step.
Their comparison detects when traversal stops progressing; a file-existence check allows early return.
\end{minipage}

\par\bigskip\noindent
\begin{minipage}{\linewidth}
\textbf{Computing a median with fallback handling.}\hfill
{\small\href{https://github.com/libxsmm/libxsmm/blob/5cadf3e2f8c5169a7d3dbe87a1b094686b0c8c29/scripts/libxsmm_utilities.py}{LIBXSMM}}
\begin{lstlisting}[style=realworldroles,
  emph={list_of_numbers},emphstyle=\bfseries\color{rworganizer}]
def median(list_of_numbers, fallback = None, average = True):
    size = len(list_of_numbers)
    if (0 < size):
        # TODO: use nth element
        list_of_numbers.sort()
        size2 = int(size / 2)
        if (average and 0 == (size - size2 * 2)):
            medval = int(0.5 * (list_of_numbers[size2-1] + list_of_numbers[size2]) + 0.5)
        else:
            medval = list_of_numbers[size2]
        if (None != fallback):
            result = min(medval, fallback)
        else:
            result = medval
    elif (None != fallback):
        result = fallback
    else:
        sys.tracebacklimit = 0
        raise ValueError("median: empty list!")
    return result
\end{lstlisting}
\small
The \emph{organizer} \textcolor{rworganizer}{\texttt{list\_of\_numbers}} sorts the elements; subsequent computation only reads the reordered list.
The function also handles even-sized input, a fallback value, and empty input.
\end{minipage}
\clearpage
\endgroup

\clearpage
\section{Corpus sizes in prior variable-role studies}
\label{app:prior-dataset-sizes}

Table~\ref{tab:prior-dataset-sizes} expands Table~\ref{tab:variable_role_extraction},
distinguishing source-code pools from labeled examples in each study's reported units.

\begingroup
\small
\setlength{\tabcolsep}{4pt}
\renewcommand{\arraystretch}{1.10}
\setlength{\LTleft}{0pt}
\setlength{\LTright}{0pt}
\setlength{\LTcapwidth}{\textwidth}
\begin{longtable}{@{}>{\raggedright\arraybackslash}p{0.20\textwidth}>{\raggedright\arraybackslash}p{0.43\textwidth}>{\raggedright\arraybackslash}p{\dimexpr0.37\textwidth-16pt\relax}@{}}
\caption{Corpus sizes and annotation scope for Table~\ref{tab:variable_role_extraction}.
Totals combine reported splits where available. NR: not reported, not zero.
Units and annotation quality are not directly comparable.}
\label{tab:prior-dataset-sizes}\\
\toprule
\textbf{Study / coverage} & \textbf{Corpus and labeled examples} & \textbf{Scope and qualifications} \\
\midrule
\endfirsthead
\multicolumn{3}{@{}l}{\tablename~\thetable\ (continued)} \\
\toprule
\textbf{Study / coverage} & \textbf{Corpus and labeled examples} & \textbf{Scope and qualifications} \\
\midrule
\endhead
\midrule
\multicolumn{3}{r@{}}{\emph{Continued on next page}} \\
\endfoot
\bottomrule
\endlastfoot

\citet{gerdt2004automatic}\par
Pascal; 3 roles &
Prototype program and variable counts: NR. &
Fixed Value, Stepper, and Most-Recent Holder, selected using an earlier textbook survey. \\
\addlinespace[5pt]

\citet{bishop2005assessing}\par
Java; role checking &
17 canonical teaching programs; variable count: NR. &
Checks supplied role assignments in procedural Java. \\
\addlinespace[5pt]

\citet{gerdt2006web}\par
Pascal; 9 roles &
Learning: 43 programs, 175 variables.
Evaluation: 44 programs, 130 variables.
Total: 87 programs, 305 variables. &
Filtered, human-annotated textbook sample; arrays omitted and role-changing variables split by role. \\
\addlinespace[5pt]

\citet{demyanova2013variableRoles}\par
C; 14 custom roles &
5.2 KLOC of cBench for role design.
Downstream file and variable totals: NR. &
Custom roles evaluated on SV-COMP file-category prediction, not role-label accuracy. \\
\addlinespace[5pt]

\citet{lui2020recognition}\par
C; 4 classes &
110 educational programs; 8,907 labeled trace cases. &
Manually annotated programs; cases are not stated to be distinct declarations.
One-Way Flag and Constant are merged. \\
\addlinespace[5pt]

\citet{hussain2023roles}\par
Java; 2 roles &
14,022,318 methods; 1,438,035 role-augmented variables
(624,701 steppers; 813,334 walkers). &
Automatic syntax rules; pool includes methods without either role.
Evaluated on method-name prediction. \\
\addlinespace[5pt]

\textbf{Ours}\par
Five languages; 11 roles &
126,926 files; 1,737,758 distinct role-annotated variables
(37,006 steppers; 41,686 walkers).
Full counts: Appendix~\ref{app:dataset}. &
Automatic CPG labels; identical sources and matching declarations merged.
Unlabeled variables can still have roles. \\
\end{longtable}
\endgroup

\paragraph{Comparability.}
Programs, methods, and files are different source units, as are trace cases, declarations, and variable--role memberships.
Counts do not imply matched deduplication, independent examples, or comparable holdouts.
Role inventories and recognition rules also differ: \citet{gerdt2006web} include Transformation but have no Temporary evaluation examples.
Automatic annotations and downstream task performance do not establish role-label accuracy.

\paragraph{Coverage.}
Our eleven-role, five-language corpus has more role-annotated variables than \citet{hussain2023roles}, but fewer steppers and walkers (Table~\ref{tab:prior-dataset-sizes}).
Coverage is uneven: Fixed Value accounts for 74.4\% of role memberships, and Ruby for 53 files.
Broader coverage does not imply more examples of each role or higher annotation quality.

\clearpage
\section{Variable-role extraction details}
\label{app:extraction}
\label{app:role-contracts}

Section~\ref{sec:ai-use} describes how coding agents assisted implementation of these author-defined contracts.

\begingroup
\setlength{\abovedisplayskip}{4pt}
\setlength{\belowdisplayskip}{4pt}
\setlength{\abovedisplayshortskip}{3pt}
\setlength{\belowdisplayshortskip}{3pt}
\newcommand{\ERp}[1]{\mathsf{#1}}
\newcommand{\ERcard}[4]{%
  \par\medskip\noindent\begin{minipage}{\linewidth}
  \textbf{#1}\par\smallskip
  #2\par\smallskip
  {\raggedright\emph{Key:} #3\par}\smallskip
  \emph{Pattern:} \texttt{#4}
  \end{minipage}\par}
\newenvironment{ERcode}{\begin{tabular}{@{}r@{\quad}l@{}}}{\end{tabular}}
\newcommand{\ERsetup}{%
\noindent $v$: declaration identity; $W$: binding writes; $I$: initialization events; $U=W\setminus I$; $M$: observed state mutations; $F$: raw Fixed Values.
Initialization partitions are role-specific; named checks require supporting source/CPG evidence.
CFG witnesses are bounded; missing labels do not prove semantic absence.\par\smallskip
{\raggedright\noindent\emph{Supplement:} \path{src/role_extractor/scala/}: \path{facts/} builds evidence, \path{role_predicates.sc} classifies, \path{annotation_emission.sc} resolves labels.
Frontend details and exact algorithms reside in this executable specification.\par}}

\newcommand{\ERfixedkeys}{%
$Q$: verified non-text collection; $C$: collection decision (third page).
$J$: verified initialization, at most one per path (parameter entry; local initializer/exhaustive alternatives; member constructors).
$B$: no later/external rebinding.
$A$: plain, non-self-dependent writes.
$Z$: no observed mutation.
$O$: no loop initialization/rebinding.}
\newcommand{\ERfixedexample}{limit = input\_limit; use(limit)}

\newcommand{\ERtempkeys}{%
$L$: genuine callable-local; $K_T$: complete read associations.
$D$: direct plain non-self assignment; RHS call/known-variable token, excluding strings/keywords and import/definition artifacts.
$R_w$: reaching-definition reads with redefinition-free CFG paths.
$d$: line difference; same-line event order; otherwise CFG instruction distance.
$H$: removing writes disconnects loop-entry/back-edge paths to reads.}
\newcommand{\ERtempexample}{tmp = parse(item); emit(tmp)}

\newcommand{\ERflagkeys}{%
$J_b$: complete canonical identity/partition, Boolean-literal initialization $b$ (also 0/1).
$T_b$: complete direct write of $\neg b$, or $v\lor e$ from false / $v\land e$ from true; symmetric operands allowed.
$e$: Boolean literal/reference/comparison or supported logical composition.
$N_b$: terminal assignment or non-inert operand; structural compatibility, not guard satisfiability.}
\newcommand{\ERflagexample}{found = false; ...; found = found || match}

\newcommand{\ERstepkeys}{%
$N$: implicit numeric iterators.
$P$: complete repeated progression; every later write a direct self-update with unary $+/-$ or binary $+,-,\times,/$; exclude $\text{fixed}-v$ and proved Walkers.
$D_v$: non-self update dependencies.
$G(v,F,S)$: complete supported guards; dependencies in $S$ or fixed inputs with predictable seeds (parameters allowed).}
\newcommand{\ERstepexample}{i = 0; while i < limit: i += stride}

\newcommand{\ERwalkerkeys}{%
$E$: implicit element iterator, excluding numeric iterators.
$K_W$: complete navigation and initialization partition.
$V$: previous-position dependence; loop context; access/control use; resolved member, fixed-collection successor index, supported navigation call, or cursor increment/decrement with direct cursor use.
An ordinary array index alone is insufficient.}
\newcommand{\ERwalkerexample}{while p: use(p); p = p.next}

\newcommand{\ERrecentkeys}{%
$K_R$: complete acquisition/correction coverage of $U$, $M=\varnothing$, no holder-dependent acquisition guard.
$A_v$: acquisitions (call, collection access, iterator projection).
$L_R$: non-member loop/proved-iterator context.
$R$: read outside own writes via one unambiguous stable-input correction chain; limits: depth 64, nodes 10,000, corrections 5, branches 128.
$P_R$: same-owner member acquisitions in ordinary methods plus an owner use; no cross-call ordering proof.}
\newcommand{\ERrecentexample}{while more: current = read(); consume(current)}

\newcommand{\ERfollowerkeys}{%
$K_F$: complete, nonempty reassignment set $U_F$.
$A_s$: uniquely resolved canonical declaration/member/fixed-index master; RHS $s$, $s+k$, $k+s$, or $s-k$; $k$ literal/raw Fixed Value.
$Y_s$: every local loop has capture $\to$ master evolution $\to$ follower use $\to$ capture; members require ordered capture/evolution per writing method and same-owner use.
$X$: self-dependence, mutation, follower-controlled write, or follower-dependent master evolution.}
\newcommand{\ERfollowerexample}{prev = current; current = next(current); use(prev)}

\newcommand{\ERwantedkeys}{%
$J_S$: compatible seeds, nonempty loop-selection epochs, one direction.
$S$: guarded holder--candidate comparison, min/max, or null-seeded first-satisfying candidate with structural exit.
$K_S$: all writes explained (seed/selection/unset fallback), no mutation.
$B_h$: nonempty later writes all coupled to one primary $h$ by a unique guard in one context; nonempty fixed/stepper/walker dependencies, later read, no mutation.}
\newcommand{\ERwantedexample}{if score > best: best = score; winner = i}

\newcommand{\ERgatherkeys}{%
$K_G$: admitted numeric non-collection; complete phase coverage of $U$.
$J_G$: parameter, stable numeric seed, or supported first contribution.
$\mathcal A$: nonempty loop spans split by resets.
$\ERp{SelfUpdate}$: uses the previous value.
$D$: indexed/member/dereferenced data or a non-fixed/non-stepper dependency; calls through argument data.
$T$: numeric/fixed resets; arithmetic transforms with numeric/fixed/stepper inputs; at most one stable arithmetic finalization per method outside loops.}
\newcommand{\ERgatherexample}{score *= 5; score += points[item]}

\newcommand{\ERcontainerkeys}{%
$Q$: verified non-text collection; $H_c$: source-backed target, excluding synthetic/function bindings.
$E_c$: changes after the identified construction phase: rebinding, updates, add/remove/clear, element/descendant writes, exact-target mutations.
$\Pi$: proved permutations; $P_c$: event IDs they consume.
Alias-only effects do not establish mutation.}
\newcommand{\ERcontainerexample}{items.append(x); items.sort()}

\newcommand{\ERorganizerkeys}{%
$J_c$: verified, complete collection initialization.
$\Pi$: tuple swap; local dataflow swap; exact same-collection swap; known in-place sort/reverse/shuffle/rotate.
Manual swaps require reaching-definition/CFG witnesses without transformation; ambiguous paths, indexes, or aliases fail.
$K_c$: complete collection lifecycle.}
\newcommand{\ERorganizerexample}{items.reverse()}

\newcommand{\ERemitkeys}{%
\noindent\emph{Emission:} retain raw multilabel results subject to exclusions above.
Suppress Temporary if any role other than Fixed Value co-occurs; otherwise Temporary suppresses Fixed Value.
Member subjects omit Temporary, Gatherer, and Most Wanted Holder.
Raw Fixed Values remain available to dependency checks.
Emit identity, labels, metadata, and exact source views; graph/alias/interprocedural limits follow the supplementary implementation.}

\subsection*{Extraction pseudocode \hfill 1/3}
\label{app:extract-pseudocode}
\ERsetup

\ERcard{Fixed Value}{%
\begin{ERcode}
1 & \textbf{if} $Q(v)$: \textbf{return} $C(v)=\ERp{Fixed}$\\
2 & \textbf{require} $J(v)$ and $B(v)$\\
3 & \textbf{reject if} any write violates $A$, or $Z(v)$ or $O(v)$ fails\\
4 & \textbf{accept}
\end{ERcode}}{\ERfixedkeys}{\ERfixedexample}

\ERcard{Temporary (raw)}{%
\begin{ERcode}
1 & \textbf{require} $L(v)$, $K_T(v)$, $W\ne\varnothing$\\
2 & \textbf{for each} $w\in W$: \textbf{require} $D(w)$ and $1\leq |R_w|\leq5$\\
3 & \phantom{\textbf{for each}} \textbf{require} $0\leq d(w,r)\leq5$ for every $r\in R_w$\\
4 & \textbf{return} $H(v)$
\end{ERcode}}{\ERtempkeys}{\ERtempexample}

\ERcard{One-Way Flag}{%
\begin{ERcode}
1 & $b\gets$ initial Boolean value; \textbf{require} $J_b(v)$, $U\ne\varnothing$, $M=\varnothing$\\
2 & \textbf{for each} $w\in U$: \textbf{require} $T_b(w)$\\
3 & \textbf{return} some $w\in U$ satisfies $N_b(w)$
\end{ERcode}}{\ERflagkeys}{\ERflagexample}

\ERcard{Stepper}{%
\begin{ERcode}
1 & $S\gets N$\\
2 & \textbf{repeat} $S\gets S\cup\{v:P(v),\ D_v\subseteq F\cup S,\ G(v,F,S)\}$\\
3 & \textbf{until} $S$ stops changing; \textbf{return} $v\in S$
\end{ERcode}}{\ERstepkeys}{\ERstepexample}

\clearpage
\subsection*{Extraction pseudocode \hfill 2/3}

\ERcard{Walker}{%
\begin{ERcode}
1 & \textbf{if} $E(v)$: \textbf{accept}\\
2 & \textbf{require} $K_W(v)$ and $U\ne\varnothing$\\
3 & \textbf{for each} $w\in U$: \textbf{require} $V(w)$\\
4 & \textbf{accept}
\end{ERcode}}{\ERwalkerkeys}{\ERwalkerexample}

\ERcard{Most Recent Holder}{%
\begin{ERcode}
1 & \textbf{require} $K_R(v)$ and $A_v\ne\varnothing$\\
2 & \textbf{reject if} Walker, Follower, or Most Wanted Holder\\
3 & \textbf{if} member: \textbf{return} $P_R(v)$\\
4 & \textbf{return} $L_R(v)$ and every $a\in A_v$ has witness $R(a)$
\end{ERcode}}{\ERrecentkeys}{\ERrecentexample}

\ERcard{Follower}{%
\begin{ERcode}
1 & \textbf{require} $K_F(v)$; \textbf{reject if} $X(v)$\\
2 & $s\gets$ unique master shared by every non-initial assignment\\
3 & \textbf{require} $A_s(w)$ for every reassignment $w$\\
4 & \textbf{return} $Y_s(v)$
\end{ERcode}}{\ERfollowerkeys}{\ERfollowerexample}

\ERcard{Most Wanted Holder}{%
\begin{ERcode}
1 & $\ERp{Primary}(v)\gets J_S(v)$ and $S(v)$ and $K_S(v)$\\
2 & \textbf{clear primary status if} Stepper, Walker, Gatherer, or Follower\\
3 & \textbf{accept if} $\ERp{Primary}(v)$\\
4 & \textbf{otherwise accept if} some primary $h$ satisfies $B_h(v)$
\end{ERcode}}{\ERwantedkeys}{\ERwantedexample}

\clearpage
\subsection*{Extraction pseudocode \hfill 3/3}
\label{app:collection-rules}

\ERcard{Gatherer}{%
\begin{ERcode}
1 & \textbf{require} $K_G(v)$, $J_G(v)$, $\mathcal A_v\ne\varnothing$\\
2 & \textbf{reject if} Stepper or Walker\\
3 & \textbf{for each} span $a\in\mathcal A_v$: \textbf{require} a self-update with $D$\\
4 & \textbf{require} all resets, other transforms, and finalizations satisfy $T$\\
5 & \textbf{accept}
\end{ERcode}}{\ERgatherkeys}{\ERgatherexample}

\ERcard{Container}{%
\begin{ERcode}
1 & \textbf{require} $Q(v)$ and $H_c(v)$\\
2 & $E_c\gets$ post-construction collection changes\\
3 & $P_c\gets$ event IDs consumed by proved permutation spans $\Pi$\\
4 & \textbf{return} $E_c\setminus P_c\ne\varnothing$
\end{ERcode}}{\ERcontainerkeys}{\ERcontainerexample}

\ERcard{Organizer}{%
\begin{ERcode}
1 & \textbf{require} $Q(v)$, $H_c(v)$, and $J_c(v)$\\
2 & \textbf{require} $\Pi_v\ne\varnothing$\\
3 & \textbf{return} $E_c\setminus P_c=\varnothing$
\end{ERcode}}{\ERorganizerkeys}{\ERorganizerexample}

\par\medskip\noindent\textbf{Collection decision and final labels}\par\smallskip
\begin{ERcode}
1 & $C(v)\gets\ERp{None}$; \textbf{require} $Q(v)$ and $H_c(v)$\\
2 & \textbf{if} $E_c\setminus P_c\ne\varnothing$: $C(v)\gets\ERp{Container}$\\
3 & \textbf{else if} $J_c(v)$ and $\Pi_v\ne\varnothing$: $C(v)\gets\ERp{Organizer}$\\
4 & \textbf{else if} $J_c(v)$ and $K_c(v)$: $C(v)\gets\ERp{Fixed}$
\end{ERcode}
\par\smallskip\ERemitkeys

\endgroup
\clearpage

\clearpage
\section{Variable Role Extractor Human Evaluation}
\label{app:variable_extractor_human_eval}

\begin{table}[t]
\centering
\small
\setlength{\tabcolsep}{5pt}
\begin{tabular}{lrrrrr}
\toprule
& \multicolumn{2}{c}{Distribution} & \multicolumn{3}{c}{Performance} \\
\cmidrule(lr){2-3}\cmidrule(lr){4-6}
Role & Extractor & Human & Precision & Recall & F1 \\
\midrule
Container           & 30  & 27  & 0.733 & 0.815 & 0.772 \\
Fixed value         & 106 & 145 & 0.830 & 0.607 & 0.701 \\
Follower            & 20  & 22  & 0.900 & 0.818 & 0.857 \\
Gatherer            & 20  & 25  & 0.900 & 0.720 & 0.800 \\
Most-recent holder  & 21  & 30  & 0.667 & 0.467 & 0.549 \\
Most-wanted holder  & 21  & 19  & 0.905 & 1.000 & 0.950 \\
One-way flag        & 21  & 26  & 1.000 & 0.808 & 0.894 \\
Organizer           & 20  & 19  & 0.850 & 0.895 & 0.872 \\
Stepper             & 23  & 38  & 0.957 & 0.579 & 0.721 \\
Temporary           & 35  & 49  & 0.829 & 0.592 & 0.690 \\
Walker              & 22  & 28  & 1.000 & 0.786 & 0.880 \\
No role             & 324 & 232 & --    & --    & --    \\
\midrule
Micro-average       & --  & --  & 0.855 & 0.678 & 0.756 \\
\bottomrule
\end{tabular}
\caption{Human evaluation of the variable role extractor on 660
stratified samples. Distribution columns report the number of
extractor predictions and human annotations per role.
Precision, recall, and F1 are computed one-vs-rest against human
labels, excluding ``no role''. Scores characterize the selected
subset rather than corpus-wide performance.}
\label{tab:human_role_eval}
\end{table}

\paragraph{Annotation protocol.}
Because Sajaniemi's roles describe cognitive rather than strictly
syntactic properties of variables, we assess the extractor against
human annotations. Given the size of the corpus and the highly uneven
role distribution, we construct a stratified sample combining
extractor positives with two types of negatives.

For each of the 11 roles, we sample 20 variables assigned that role
by the extractor and 20 variables not assigned that role but whose
names suggest it (e.g., \emph{best} for most-wanted holder).
Rather than defining these semantic priors manually,
we identify identifier substrings that best discriminate each role
in the extracted corpus, ranked by their F1 score, and use them to
sample potential false negatives. This enriches the sample with
plausible missed instances of rare roles, such as follower, which
would be unlikely to appear in a random sample and would otherwise
make recall difficult to assess.
We additionally sample 220 variables without any extracted role,
yielding 660 distinct variables.

Variables are presented in random order, with their identifier occurrences highlighted in a code excerpt containing all variable occurrences. The annotator assigns one of the 11 roles or ``no role'', without access to extractor predictions or sampling strata.

\paragraph{Evaluation.}
We treat human annotations as reference labels and evaluate the extractor as a multi-label classifier, reporting one-vs-rest precision, recall, and F1 for each role. Table~\ref{tab:human_role_eval} summarizes the sample and human-label distributions, and reports the resulting scores. Because the sample deliberately balances roles and enriches negatives with semantic priors, these unweighted scores characterize the selected subset rather than corpus-wide performance.

\section{Dataset statistics}
\label{app:dataset}

The consolidated corpus combines the earlier role-enriched dataset with variable records from successful shards of the expanded CodeParrot annotation run.
Of 150,000 new candidate files, thirteen shards completed and two failed parsing; failed shards contribute no new annotations.
We merge byte-identical sources and identify variables by source hash, kind, name, and declaration coordinates, preserving both releases' annotation IDs.
The corpus contains 126,926 source files, 14,624,580 physical lines, and 3,525,530 unique variables across five languages, retaining the earlier Ruby data.
Of these, 1,737,758 have an extracted role and 1,787,772 have none; missing labels do not imply semantic absence of a role.

\begin{table}[!ht]
\centering
\caption{Consolidated corpus coverage by partition and language.
Role pairs count variable--role memberships; with role is the fraction of variables with at least one extracted role.}
\label{tab:dataset_coverage}
\begingroup\small\setlength{\tabcolsep}{2pt}
\begin{minipage}[t]{0.50\linewidth}\centering
\begin{tabular}{@{}lrrr@{}}
\toprule
Partition & Files & Variables & Role pairs \\
\midrule
Train & 76,155 & 2,116,300 & 1,048,604 \\
Validation & 25,386 & 705,213 & 349,496 \\
Test & 25,385 & 704,017 & 349,527 \\
\midrule
All & 126,926 & 3,525,530 & 1,747,627 \\
\bottomrule
\end{tabular}

\end{minipage}\hfill
\begin{minipage}[t]{0.48\linewidth}\centering
\begin{tabular}{@{}lrrr@{}}
\toprule
Language & Files & Variables & With role \\
\midrule
C & 32,492 & 964,660 & 42.8\% \\
C++ & 31,910 & 913,138 & 49.3\% \\
JavaScript & 29,673 & 951,288 & 45.1\% \\
Python & 32,798 & 695,422 & 64.0\% \\
Ruby & 53 & 1,022 & 69.6\% \\
\bottomrule
\end{tabular}

\end{minipage}
\endgroup
\end{table}

Five approximately equal file folds are stratified by role and language with source groups kept intact.
For the baselines, three folds train, one validates, and one tests, yielding 2,116,300/705,213/704,017 variables (Table~\ref{tab:dataset_coverage}).
The corpus includes 2,428,645 locals, 865,797 parameters, and 231,088 members.

\begin{table}[!ht]
\centering
\caption{Role support in the consolidated corpus.
Percentages use all 3,525,530 variables; test files counts files in each row.
Extracted (total) sums variable--role and file--role memberships across rows, not unique variables or files.
``No extracted role'' is not a prediction target.}
\label{tab:role_distribution}
\begingroup\small\setlength{\tabcolsep}{3pt}
\begin{tabular}{@{}lrrrrrr@{}}
\toprule
Role & Train & Valid. & Test & Total & \% & Test files \\
\midrule
Fixed value & 780,012 & 259,968 & 260,007 & 1,299,987 & 36.87 & 24,387 \\
Stepper & 22,201 & 7,403 & 7,402 & 37,006 & 1.05 & 4,079 \\
Gatherer & 2,877 & 959 & 958 & 4,794 & 0.14 & 677 \\
Most-recent holder & 6,500 & 2,166 & 2,168 & 10,834 & 0.31 & 1,300 \\
Most-wanted holder & 901 & 298 & 300 & 1,499 & 0.04 & 217 \\
One-way flag & 4,321 & 1,439 & 1,437 & 7,197 & 0.20 & 1,132 \\
Follower & 435 & 144 & 145 & 724 & 0.02 & 125 \\
Walker & 25,013 & 8,337 & 8,336 & 41,686 & 1.18 & 4,250 \\
Container & 53,230 & 17,744 & 17,738 & 88,712 & 2.52 & 7,969 \\
Organizer & 448 & 150 & 149 & 747 & 0.02 & 127 \\
Temporary & 152,666 & 50,888 & 50,887 & 254,441 & 7.22 & 15,521 \\
\midrule
Extracted (total) & 1,048,604 & 349,496 & 349,527 & 1,747,627 & 49.57 & 59,784 \\
No extracted role & 1,073,661 & 357,596 & 356,515 & 1,787,772 & 50.71 & 24,348 \\
\bottomrule
\end{tabular}

\endgroup
\end{table}

Fixed value contributes 74.4\% of the 1,747,627 memberships, while follower and organizer have 724 and 747 variables (Table~\ref{tab:role_distribution}).
There are 9,866 multi-role variables; the largest overlap is fixed value with walker, for 5,665 variables.

\paragraph{Integrity and scope.}
Exact-source matches, normalized whole-file/function matches, and earlier clone links define 122,458 indivisible source groups; none crosses folds.
This grouping is not repository holdout or exhaustive semantic deduplication; 16,784 files have parser errors.
We union existential boolean flags across duplicates and recover flags for variables with no extracted role from raw facts.
The earlier corpus was used in model development; the new split does not erase that exposure.

\clearpage
\ifnewdatasetappendices
\begingroup
\renewcommand{\thesection}{\texorpdfstring{D$^\prime$}{D'}}
\renewcommand{\theHsection}{D-new}
\addtocounter{table}{-2}
\renewcommand{\thetable}{\texorpdfstring{\arabic{table}$^\prime$}{\arabic{table}'}}
\renewcommand{\theHtable}{D-new.\arabic{table}}
\input{appendices/dataset_statistics_new}
\clearpage
\endgroup
\addtocounter{section}{-1}
\fi
\section{Metadata and lexical baselines}
\label{app:dataset-baselines}

We use the full consolidated corpus (Appendix~\ref{app:dataset}), without subsampling.
The split has 2,116,300 training, 705,213 validation, and 704,017 test variables.
For each of eleven roles, variables lacking that annotation, including those with no extracted role, are negative.
We evaluate extractor labels using original identifiers.

\paragraph{Predictors and selection.}
We use language/kind categories, six general boolean properties, and TF--IDF name character 1--4-grams or local-code token/operator 1--2-grams.
Vocabularies retain at most 12,000 name and 30,000 code features, with minimum document frequency 3.
Code windows cover up to eight occurrences with 100 source bytes per side; verified target identifiers are masked.
The combined baseline concatenates name, code, and boolean features; role-specific metadata and target-derived flags are excluded.
For each role, class-balanced L2 logistic regression selects $C\in\{0.1,1,10\}$ by validation AP and a threshold by validation F1.
Vocabularies and weights are refitted on 2,821,513 train/validation variables; thresholds are unchanged.

\begin{table}[!ht]
\centering
\caption{Macro surface-baseline performance over eleven roles.
Brackets: 95\% intervals from 500 source-group bootstrap resamples, conditional on fitted models; $\pm$: standard deviation over 500 random draws.}
\label{tab:dataset_baselines}
\begingroup\small\setlength{\tabcolsep}{4pt}
\begin{tabular}{@{}lrr@{}}
\toprule
Predictor & Macro AP & Macro F1 \\
\midrule
Binary majority & 0.045 & 0.000 \\
Most frequent role & 0.045 & 0.049 \\
Always positive & 0.045 & 0.071 \\
\midrule
Random (uniform) & $0.0451\pm0.0001$ & $0.0598\pm0.0001$ \\
Random (prior-matched) & $0.0451\pm0.0001$ & $0.0451\pm0.0004$ \\
\midrule
Language + kind & 0.100 [0.098, 0.101] & 0.112 [0.109, 0.114] \\
Boolean properties & 0.067 [0.065, 0.069] & 0.102 [0.099, 0.104] \\
Variable name & 0.142 [0.140, 0.146] & 0.213 [0.207, 0.218] \\
Local code & 0.487 [0.475, 0.503] & 0.526 [0.515, 0.537] \\
Combined & 0.533 [0.521, 0.547] & 0.553 [0.541, 0.564] \\
\bottomrule
\end{tabular}

\endgroup
\end{table}

Train/validation frequencies define binary majority (no roles) and most-frequent-role (fixed value) references; always positive predicts every role.
Uniform random scores yield positive predictions with probability $0.5$ or the train/validation role frequency.
AP integrates precision over recall increments at distinct score thresholds; F1 is $2\mathrm{TP}/(2\mathrm{TP}+\mathrm{FP}+\mathrm{FN})$ at the frozen threshold.
Constant-score AP equals test prevalence; macro averages weight roles equally.

\begin{table}[!ht]
\centering
\caption{Per-role surface predictability on the consolidated corpus.
Prior AP is test prevalence; code denotes local code, and combined adds names and boolean properties.}
\label{tab:dataset_baseline_roles}
\begingroup\small\setlength{\tabcolsep}{4pt}
\begin{tabular}{@{}lrrrrr@{}}
\toprule
Role & Prior AP & Name AP & Code AP & Combined AP & Combined F1 \\
\midrule
Fixed value & 0.3693 & 0.594 & 0.785 & 0.822 & 0.752 \\
Stepper & 0.0105 & 0.363 & 0.649 & 0.679 & 0.690 \\
Gatherer & 0.0014 & 0.075 & 0.423 & 0.479 & 0.535 \\
Most-recent holder & 0.0031 & 0.011 & 0.222 & 0.223 & 0.317 \\
Most-wanted holder & 0.0004 & 0.056 & 0.388 & 0.458 & 0.488 \\
One-way flag & 0.0020 & 0.052 & 0.516 & 0.665 & 0.658 \\
Follower & 0.0002 & 0.008 & 0.099 & 0.139 & 0.205 \\
Walker & 0.0118 & 0.114 & 0.882 & 0.893 & 0.840 \\
Container & 0.0252 & 0.115 & 0.588 & 0.635 & 0.620 \\
Organizer & 0.0002 & 0.009 & 0.272 & 0.299 & 0.378 \\
Temporary & 0.0723 & 0.170 & 0.538 & 0.569 & 0.597 \\
\bottomrule
\end{tabular}

\endgroup
\end{table}

Binary majority attains 95.5\% accuracy with zero F1.
Name, code, and combined macro AP are 0.142, 0.487, and 0.533 (Tables~\ref{tab:dataset_baselines}--\ref{tab:dataset_baseline_roles}).
These scores measure automatic-label predictability, without independent semantic validation.
Probe comparisons require matched subjects, negatives, partitions, and accessible context; these windows can include code after a causal readout.

\clearpage
\ifnewdatasetappendices
\begingroup
\renewcommand{\thesection}{\texorpdfstring{E$^\prime$}{E'}}
\renewcommand{\theHsection}{E-new}
\addtocounter{table}{-2}
\renewcommand{\thetable}{\texorpdfstring{\arabic{table}$^\prime$}{\arabic{table}'}}
\renewcommand{\theHtable}{E-new.\arabic{table}}
\input{appendices/dataset_baselines_new}
\clearpage
\endgroup
\addtocounter{section}{-1}
\fi
\section{SAE Details}
\label{app:sae-details}

\begin{table}[ht]
\centering
\caption{
Sparse autoencoder configurations used for each model.
}
\label{tab:model_sae_configurations}
\small
\setlength{\tabcolsep}{4pt}
\def\UrlBreaks{\do\/\do\_\do\-\do\.}
\begin{tabular}{@{}>{\raggedright\arraybackslash}p{0.14\linewidth}>{\raggedright\arraybackslash}p{0.30\linewidth}>{\raggedright\arraybackslash}p{0.34\linewidth}>{\raggedright\arraybackslash}p{0.13\linewidth}@{}}
\toprule
Model & SAE release & SAE ID & Layers \\
\midrule

Gemma~2~2B
& \path{gemma-scope-2b-pt-res-canonical}
& \path|layer_{l}/width_16k/canonical|
& 6, 13, 17, 22 \\

Gemma~3~4B
& \path{gemma-scope-2-4b-pt-res}
& \path|layer_{l}_width_65k_l0_medium|
& 9, 17, 22, 29 \\

Gemma~3~12B
& \path{gemma-scope-2-12b-pt-res}
& \path|layer_{l}_width_65k_l0_medium|
& 12, 24, 31, 41 \\

Llama~3.1~8B
& \path{llama_scope_lxr_32x}
& \path|l{l}r_32x|
& 8, 16, 21, 27 \\

Qwen3.5~9B
& \path{qwen-scope-3.5-9b-base-w64k-l100}
& \path|layer{l}|
& 8, 16, 21, 27 \\

\bottomrule
\end{tabular}
\end{table}

\begingroup
\makeatletter
\setlength{\@fptop}{89.8336pt}
\setlength{\@fpbot}{0pt plus 1fil}
\clearpage
\makeatother
\endgroup
\section{Additional Analyses of Role Representations}
\label{app:additional-role-analyses}
\label{sec:additional-role-analyses}

\begin{figure}[!ht]
\centering
\includegraphics[width=\columnwidth]{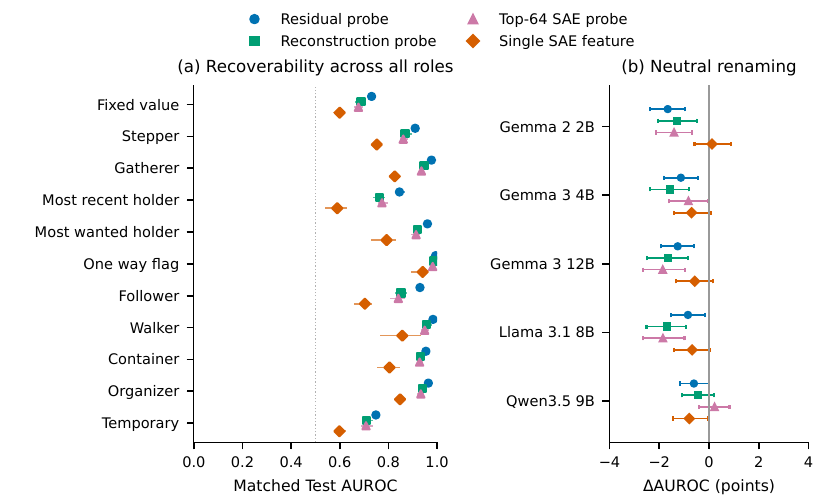}
\caption{
\textbf{Residual probes recover more role information than individual SAE features, with modest sensitivity to neutral target names.}
(a) Matched-negative Test AUROC for all eleven roles; points average five models and their four prescribed layers, and lines span the five model means.
(b) Change in AUROC after renaming both members of 100 positive/negative pairs per role, with fits and thresholds fixed.
Intervals are paired 95\% source-group bootstrap intervals; negative values indicate lower discrimination after renaming.
}
\label{fig:role_recoverability_semantic_prior}
\vspace{12pt}
\captionsetup{type=figure}
\centering
\includegraphics[width=\columnwidth]{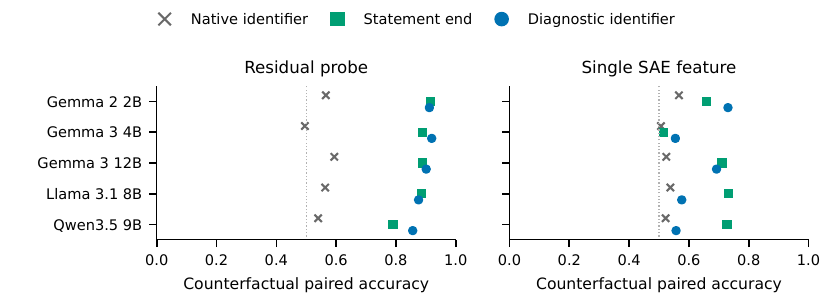}
\caption{
\textbf{Later readouts expose counterfactual changes that the native identifier often misses.}
Reviewed-counterfactual paired accuracy at the second prescribed layer of each model, averaged over all eleven roles.
Predictors are fitted separately at each readout with the same training policy and evaluation pairs; the comparison measures readout-specific decodability.
}
\label{fig:activation_localization}
\end{figure}

On matched pairs, residual probes recover automatic labels more accurately than individual SAE features, and surface cues also predict these labels.
We next test how this discrimination changes with target names, language, implementation, and the observation position.
These analyses use the fitted predictors and diagnostic readout from Section~\ref{sec:decodability}, averaging four layers except in the readout comparison.
The principal comparisons use matched-negative training; the broader sweep trains separately on matched, neighboring-role, same-context, and hard-mined negatives and evaluates all six strategies in Appendix~\ref{app:negative-strategies}.
For thresholded results, validation fixes each decision threshold by maximizing sensitivity minus false-positive rate.
We use Gemma Scope for Gemma~2~2B \citep{lieberum2024gemmascope}, Gemma Scope~2 for Gemma~3~4B and 12B \citep{mcdougall2025gemmascope2}, Llama Scope for Llama~3.1~8B \citep{he2024llamascope}, and Qwen-Scope for Qwen3.5~9B \citep{deng2026qwenscope}.

\paragraph{Reconstruction and counterfactual controls.}
To separate reconstruction loss from the restriction to one feature, we also fit ridge probes to SAE reconstructions and to the 64 training-ranked features.
Across models, their matched-negative macro AUROC reaches 0.857--0.881 and 0.853--0.878, respectively, compared with 0.906--0.913 for residual probes (Table~\ref{tab:representation-controls}).
Thus, combining features closes much of the single-feature gap.
We also test 1,698 source-reviewed role-changing counterfactual pairs from 948 source groups, excluding rejected edits and incorrect original-role labels through the LLM review described in Appendix~\ref{app:evaluation-protocol}.
Paired accuracy measures how often an original ranks above its edited negative, assigning ties half credit.
Residual paired accuracy is 0.813--0.877; reconstruction probes are 4.0--12.3 percentage points lower, with all five paired 95\% intervals excluding zero.
SAE reconstruction therefore reduces recoverability of these distinctions under the tested linear fits.
Intervals use 1,000 source-group bootstrap draws, with equal role and layer weight, conditional on the fixed split and fits.
At the same frozen validation thresholds, residual-probe false-positive rates on edited negatives are 0.684--0.713, with balanced accuracy 0.610--0.637.
These decision metrics distinguish sensitivity of score rankings from successful rejection of edited negatives.

\paragraph{Target names contribute to discrimination.}
To isolate target-name sensitivity, we select 100 Test pairs per role without consulting scores and replace every occurrence of the resolved target binding with a neutral identifier.
Both positive and negative variables are renamed, preserving their roles and the surrounding syntax; fitted choices are unchanged.
Both name conditions use the same model, SAE, and inference configuration.
Residual AUROC decreases by 0.6--1.7 percentage points across models (Figure~\ref{fig:role_recoverability_semantic_prior}b), remaining at 0.878--0.887 after renaming.
Single-feature AUROC is lower, at 0.693--0.760, with changes between $-0.8$ and $+0.1$ points.
On these same intervention pairs, name surface features fall from AUROC 0.602 to 0.492, whereas code surface features change from 0.740 to 0.723.
The residual probe preserves more discrimination than either surface baseline, although its frozen decisions still change for 10.8\% of positives and 12.4\% of negatives.
Its balanced accuracy decreases from 0.830 to 0.809, distinguishing stable ranking from stable thresholded decisions.
The edit preserves other identifiers, comments, and implementation cues, so it measures sensitivity to the target name.

\begin{figure*}[t]
\centering
\includegraphics[width=\textwidth]{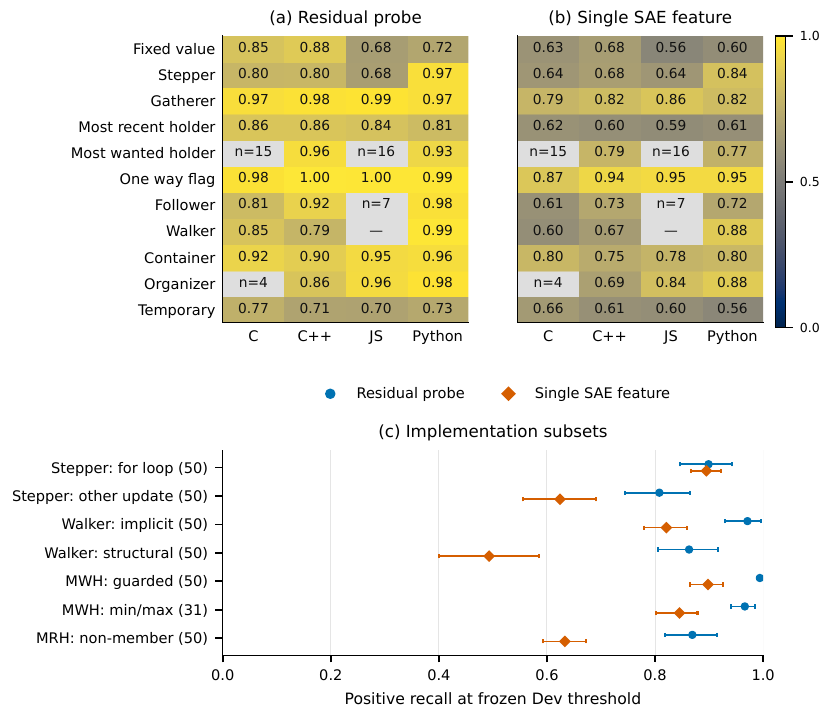}
\caption{
\textbf{Language-conditioned discrimination and implementation recall differ across roles and predictors.}
(a--b) AUROC averaged over five models and four prescribed layers, targeting 50 matched pairs per role/language.
Gray cells show cohorts below 20 pairs or ten positive source groups; a dash marks an unsupported cohort.
(c) Recall on metadata-defined positive subsets, with sample counts and 95\% source-group bootstrap intervals.
Circles denote residual probes and diamonds single SAE features; MWH and MRH abbreviate most-wanted and most-recent holder.
Implicit walkers are all Python, whereas 49 of 50 structural walkers are C/C++, so their pooled contrast also changes language.
}
\label{fig:language_implementation_dependence}
\end{figure*}

\paragraph{Language and implementation condition the measured signal.}
We supplement Test coverage toward 50 matched pairs per supported role/language and examine metadata-defined realizations (Figure~\ref{fig:language_implementation_dependence}).
The fits retain their original multilingual training and validation populations, and evaluation covers C, C++, JavaScript and Python.
Gatherer and one-way flag have high residual AUROC across all four languages, while stepper ranges from 0.682 in JavaScript to 0.972 in Python.
Ranking and thresholded decisions also differ across implementations: single-feature stepper AUROC is similar for for-loop and other updates (0.728 versus 0.757), but recall drops from 0.895 to 0.624 at the validation thresholds.
Within Python, its recall also falls from 0.904 to 0.750, on 23 and 30 positives respectively.
Most-wanted-holder recall is high for both guarded replacement and min/max selection, especially for residual probes (0.994 and 0.966).
Walker differences are entangled with language in the selected subtypes, and member-expression most-recent holders lack a supported diagnostic readout.
These distinctions constrain which apparent subtype effects can be attributed to implementation.

\paragraph{The observation point determines which edits are visible.}
Causal attention makes the readout position consequential: an assignment's identifier can precede the expression that determines its role.
At the second prescribed layer, residual paired accuracy on reviewed edits is 0.495--0.594 at the native identifier and 0.856--0.920 at the diagnostic identifier (Figure~\ref{fig:activation_localization}).
Statement-end readouts recover much of this sensitivity, with residual paired accuracy of 0.790--0.915 on the unmodified input.
These separately fitted readouts measure position-specific decodability: later positions expose source text that can be unavailable at the native identifier, and the diagnostic reference also changes the input.
The resulting predictive associations motivate the intervention experiments in Section~\ref{sec:causal-influence}.

\clearpage
\section{Evaluation populations, review, and selection}
\label{app:evaluation-protocol}

The full annotation corpus, representation benchmark, and generation tasks answer different questions.
Table~\ref{tab:evaluation-protocol} identifies their populations and selection units.

\begin{table}[!ht]
\centering
\small
\setlength{\tabcolsep}{4pt}
\begin{tabular}{>{\raggedright\arraybackslash}p{0.19\linewidth}>{\raggedright\arraybackslash}p{0.37\linewidth}>{\raggedright\arraybackslash}p{0.37\linewidth}}
\toprule
Comparison & Population and split & Selection and outcome \\
\midrule
Corpus baselines & 3,525,530 variables; three training folds, one validation fold, one Test fold & Automatic annotations; metadata and lexical prediction on the full consolidated corpus. \\
\addlinespace
Representation fits & 5,321 training and 9,547 validation pairs; source groups separated from Test & Training fixes normalization and feature ranking; validation chooses regularization, single feature, and threshold. \\
\addlinespace
Matched Test & 4,238 positive/negative pairs after exclusions; four languages & Shared examples across predictors; AP and AUROC measure rankings against automatic labels. \\
\addlinespace
Counterfactual Test & 1,698 accepted pairs from 948 source groups & Static source review of original and edited behavior; paired ranking and frozen-threshold decisions. \\
\addlinespace
Neutral-name Test & 100 positive/negative pairs per role & Both bindings renamed, with fits fixed; paired change in AUROC and binary decisions. \\
\addlinespace
Feature-scale steering & Union of 121 MBPP tasks; role-specific cohorts & Eleven SAE targets each for Gemma~2B/4B and nine for 12B; ten same-layer random features per target. \\
\addlinespace
Normalized steering & Twelve calibration and twelve Test tasks per role; eleven completed model--role cells & Shared probe/SAE dose under a correctness constraint; random residual and SAE controls. \\
\bottomrule
\end{tabular}
\caption{Evaluation populations and fitted choices.
Pair and target--task counts can reuse bindings or tasks; uncertainty resamples the corresponding source groups or unique tasks.
The separate Gemma~12B stepper pilot uses 48 calibration and 144 Test tasks (Appendix~\ref{app:causal-consolidated}).
}
\label{tab:evaluation-protocol}
\end{table}

\paragraph{Frozen representation cohort.}
The representation experiments retain an earlier prepared cohort and its source/clone-group assignments; the expanded corpus's full-variable baselines are a separate evaluation.
One rotation assigns folds 2/3/4 to training, fold 1 to validation, and fold 0 to Test.
Training and validation retain their original multilingual populations.
The original matched Test cohort contains 4,245 pairs; excluding Ruby pairs and the two source-role labels rejected by subsequent review leaves 4,238.
The common tokenization support uses a 4,096-token ceiling.
All methods and model--SAE systems share eligible pairs within a comparison; the four prescribed layers are averaged without Test-based layer selection.

\paragraph{Negative selection.}
Matched negatives share the positive's partition, programming language, and declaration kind, and lack the target annotation.
Retrieval ranks candidates using standardized context length, occurrence count, name length and shape, counts of loops, branches, assignments and calls, parser status, and identifier spelling similarity.
Scaling is fitted on training subjects; candidates from the positive's source group are excluded, with limits on negative reuse.
Neighboring-role selection additionally requires a specified competing role; same-context selection chooses another resolved binding in the enclosing context.
Hard-negative selection uses a logistic classifier on target-masked code, fitted with source-group-held-out folds within training.
The broader sweep varies the training-negative policy, retaining matched-negative validation; the main comparisons use matched training throughout.

\paragraph{Predictor fitting.}
Residual, reconstruction, and top-64 probes use ridge regression with training-only centering and coordinate scaling.
The coordinate standard-deviation floor is $10^{-6}$, and standardized vectors are divided by the square root of their dimension.
Validation AUROC selects the penalty from $\{10^{-6},10^{-4},10^{-2},1,100\}$.
Training AUROC ranks SAE coordinates by $\max(\mathrm{AUROC},1-\mathrm{AUROC})$ and fixes their predictive polarity.
Validation selects one of the top 64 coordinates for the single-feature predictor; the multi-feature probe uses those same 64 candidates.
For each predictor, validation selects the threshold maximizing sensitivity minus false-positive rate.
The matched name and visible-prefix baselines fit TF--IDF/logistic models on the same training pairs and use validation for regularization and threshold selection.
Name features use character 2--5-grams and a 12,000-feature cap; code prefixes use 3--5-grams and a 30,000-feature cap.
Both use minimum document frequency two, case-sensitive sublinear term frequencies, and validation-selected $C\in\{0.1,1,10\}$.
The code-prefix input ends at the diagnostic identifier; the full-corpus lexical windows in Appendix~\ref{app:dataset-baselines} use a different context policy.

\paragraph{Counterfactual review and exclusions.}
Counterfactuals are generated from eligible source patterns and reviewed through LLM assessments of the enclosing binding context and exact edit.
Review records specify whether the original has the target role, whether the edit removes it, source-level validity, and assumptions about external inputs or calls.
The rubric asks reviewers to inspect the binding lifetime and all writes, including mutation and aliasing, and to flag unresolved dependencies or role ambiguity.
Source annotations and proposed edits are supplied as fallible evidence; the judgments are not blinded to those proposals.
The final review uses GPT-5.6 Terra source reviewers to resolve 333 previously pending cases: 330 edits are accepted and three rejected.
One rejected edit preserves the target role; two cases have incorrect original-role labels.
Those two source-role labels are excluded from affected test panels; training and validation fits are unchanged.
The resulting 1,698 accepted pairs include earlier reviewed cases and the final accepted additions after evaluation exclusions.
These static LLM judgments are distinct from independent human adjudication or execution-based verification; the acceptance fraction of the selected queue is not an estimate of corpus-wide annotation accuracy.
Review uncertainty and inter-annotator agreement are unmeasured.

\paragraph{Ranking, decisions, and uncertainty.}
Paired counterfactual accuracy is $\mathbb{E}[\mathbf{1}(s_{\mathrm{original}}>s_{\mathrm{edit}})+\tfrac12\mathbf{1}(s_{\mathrm{original}}=s_{\mathrm{edit}})]$.
AUROC instead compares all positive/negative score combinations within a role--layer cell.
Balanced accuracy is the mean of positive recall and negative recall at the frozen threshold.
Model summaries average roles and layers equally.
Paired 95\% intervals use 1,000 joint source-group bootstrap resamples, retaining related examples and methods together; they condition on the split, fits, and review decisions.
Layer ranges in Figure~\ref{fig:role_ap} and random-feature ranges in Figure~\ref{fig:causal_scores_and_pass_at_1} describe heterogeneity, not sampling intervals.

\paragraph{Neutral names and readout controls.}
Test-pair selection does not consult predictor scores.
Neutral identifiers are derived from a role-independent binding hash; edits replace all annotated occurrences of the resolved binding, with source hashes, syntax shape, and non-target leaf bytes checked.
Original and renamed comparisons use matched execution settings; stored scores are reused only after overlap checks.
Readout controls fit separate predictors at the native identifier, statement end, and inserted identifier at each model's second prescribed layer.
They compare position-specific decodability, with differences in accessible context and, for the inserted reference, input text.

\paragraph{Steering selection and provenance.}
The representative features in Figure~\ref{fig:causal_scores_and_pass_at_1} are a descriptive selection from the larger feature-scale experiment.
The archived selection does not establish that these five representatives were chosen before intervention outcomes were inspected; aggregate conclusions therefore use all selected targets within each protocol.
Addition strength is calibrated on annotation induction for the feature-scale experiments, while the normalized experiments calibrate a joint induction-and-passing objective subject to a correctness constraint.
Appendices~\ref{app:matched-steering} and~\ref{app:causal-consolidated} report their different perturbation scopes, task overlap, denominators, and unresolved historical scale provenance.

\clearpage
\section{Behavioral role recognition and prompt sensitivity}
\label{app:behavioral-experiments}

We evaluate Llama~3.2~1B/3B and 3.1~8B, Gemma~3~1B/4B/12B/27B, Qwen3~1.7B/4B/8B/14B Base, and Qwen3.5~9B Base.
These twelve pretrained checkpoints share the protocol but vary in size, architecture, and training.

\paragraph{Evaluation and scoring.}
The same 9,553 matched positive/negative development pairs cover eleven roles and five languages: 17,770 distinct role/variable queries from 16,837 variables and 6,996 source groups.
Labels are automatic annotations; a negative lacks the target annotation.
Prompts include the role definition, full enclosing code with comments, and target identifier and location, asking for a judgment from initialization, updates, and use while allowing other roles to apply.
Each role has one fixed training-only demonstration set (seed 17), with nested, class-balanced prefixes of 0, 2, 4, or 8 examples shared across models.
Inference uses BF16 without quantization, chat templates, or source truncation.
At the final \texttt{Answer:} position, $s=\log P(\texttt{Yes})-\log P(\texttt{No})$ predicts positive when $s\geq0$; average precision (AP) ranks the unrounded scores.
We restore original pair multiplicities and average roles equally: constant predictions attain 50\% accuracy and constant scores attain 50\% AP.

\begin{figure}[!ht]
\centering
\includegraphics[width=\linewidth]{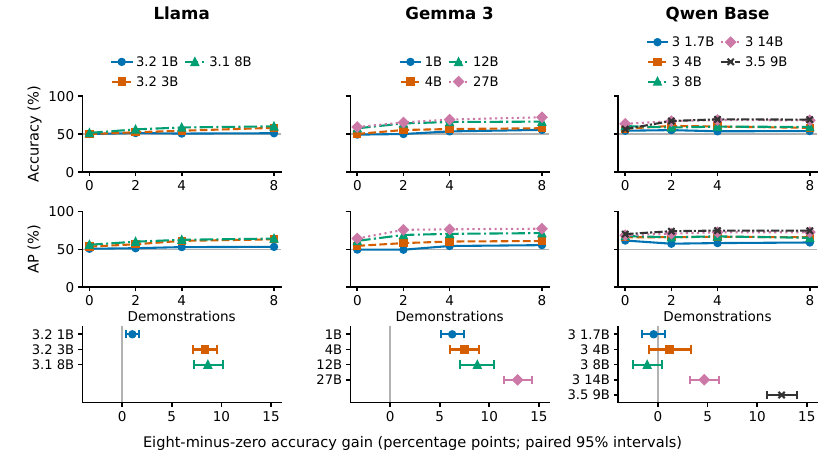}
\caption{Demonstration gains differ across checkpoints.
Top and middle: role-macro accuracy and AP on identical development pairs; gray lines mark 50\%.
Bottom: eight-minus-zero accuracy gains with paired 95\% source-group bootstrap intervals (2,000 draws), conditional on the cohort, pair weights, and one demonstration set.
Intervals are exploratory and unadjusted for multiple comparisons.
Model labels specify version and nominal size within each family.}
\label{fig:behavioral-macro}
\end{figure}

\paragraph{Recognition across checkpoints.}
Gemma~3~27B has the highest eight-shot macro accuracy/AP (72.1/77.2\%; Figure~\ref{fig:behavioral-macro}); its accuracy exceeds Gemma~3~12B by 5.8 points (paired 95\% interval: 4.6--6.9).
Qwen3~14B reaches 68.1/72.6\%, exceeding Qwen3~8B accuracy by 8.7 points (7.5--9.9).
Eight-shot accuracy rises with tested size within each family, but demonstration gains vary: the eight-minus-zero accuracy intervals include zero for Qwen3~1.7B/4B/8B, and their AP does not improve.
Llama~3.2~1B gains 1.0 accuracy point (0.3--1.7), with accuracy near chance.

\paragraph{Taxonomy-free descriptions.}
Prompts request at most three sentences on initialization, updates, and use, without role names or examples.
Training-only format calibration reduces descriptions reaching the generation cap from 27/66 to 6/66 across three checkpoints; we fix this format for all models.
Cap rates range from 0\% to 78.2\% across the expanded model set.
We assess fidelity qualitatively: Gemma~3~27B invents a prior \texttt{None} initialization for \texttt{exit\_code}, while Qwen3~14B identifies the shown assignment.

\clearpage
\subsection{Role differences and free generation}
\label{app:behavioral-prompt-formats}

\begin{figure}[!ht]
\centering
\includegraphics[width=\linewidth]{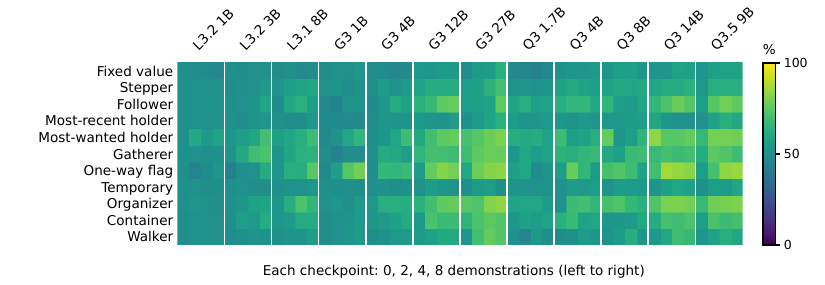}
\caption{Role accuracy varies within these aggregate trends.
Each checkpoint has four columns for 0/2/4/8 demonstrations: all 528 combinations, with original pair weights and a shared scale.
L/G/Q denote Llama/Gemma/Qwen; numerals specify version and size.}
\label{fig:behavioral-roles}
\end{figure}

\paragraph{Role and answer dependence.}
Fixed value, most-recent holder, and temporary are weaker than one-way flag and most-wanted holder (Figure~\ref{fig:behavioral-roles}).
Greedy generation evaluates 100 distinct development cases per role (50 positive, 50 negative), shared across models: 1,100 cases and 39,600 outputs across descriptions and zero/eight-shot answers.
The fixed seeded panel includes the original six cases per role; cases receive equal weight and share source groups.
Leading Yes/No answers are valid; invalid outputs count as incorrect, with a 512-new-token budget.
Three Llama~3.1~8B shards use A100/GB10 GPUs; the others use H200s, with matching software.

\begin{figure}[!ht]
\centering
\includegraphics[width=\linewidth]{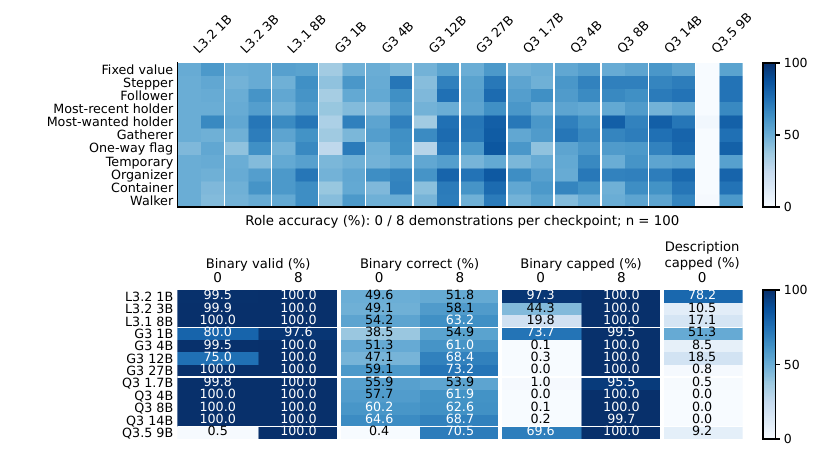}
\caption{Valid answer beginnings coexist with frequent budget exhaustion.
Top: all 264 role/model/shot accuracies, each on 100 cases.
Bottom: aggregate percentages over 1,100 cases per checkpoint; columns give demonstration counts.
Invalid answers count as errors; description caps measure stopping behavior, not semantic accuracy.}
\label{fig:behavioral-generation}
\end{figure}

Eight-shot answer validity is 97.6--100\%, yet 95.5--100\% of outputs reach the cap (Figure~\ref{fig:behavioral-generation}).
Gemma~3~27B reaches 73.2\% free-answer accuracy.
Qwen3 Base checkpoints yield 99.8--100\% valid zero-shot starts, compared with 6/1,100 for Qwen3.5~9B despite its 70.1\% zero-shot AP under constrained scoring.

\clearpage
\section{Extended related work}
\label{app:related-work}
\begingroup
\small
\setlength{\tabcolsep}{4pt}
\renewcommand{\arraystretch}{1.10}
\setlength{\LTleft}{0pt}
\setlength{\LTright}{0pt}
\setlength{\LTcapwidth}{\textwidth}
\begin{longtable}{@{}>{\raggedright\arraybackslash}p{0.27\textwidth}>{\raggedright\arraybackslash}p{0.25\textwidth}>{\raggedright\arraybackslash}p{\dimexpr0.48\textwidth-16pt\relax}@{}}
\caption{Extended comparison of work on code-language-model behavior and representations.
Groups follow Section~\ref{sec:related}, with a final group for explanation methods and human comparison; studies are placed by their primary emphasis.
The analysis column distinguishes input perturbations, representation readouts, and activation interventions.
AST: abstract syntax tree; SAE: sparse autoencoder.
${}^{\dagger}$: cited as a preprint or technical note.}
\label{tab:extended-related-work}\\
\toprule
\textbf{Study} & \textbf{Analysis and target} & \textbf{Key finding} \\
\midrule
\endfirsthead
\multicolumn{3}{@{}l}{\tablename~\thetable\ (continued)} \\
\toprule
\textbf{Study} & \textbf{Analysis and target} & \textbf{Key finding} \\
\midrule
\endhead
\midrule
\multicolumn{3}{r@{}}{\emph{Continued on next page}} \\
\endfoot
\bottomrule
\endlastfoot

\multicolumn{3}{@{}l}{\textbf{Behavioral robustness and semantic priors}} \\*
\addlinespace[2pt]
\citet{yang2022natural} & Input edits: semantics-preserving names & Plausible renaming induces failures; adversarial training improves robustness. \\
\citet{gao2023identifiers} & Input counterfactuals: names & Names provide semantics and shortcuts; counterfactual inference improves robustness. \\
\citet{wang2023recode} & Input edits: ReCode & Generation degrades under identifier, syntax, formatting, and docstring perturbations. \\
\citet{ahmed2023understanding}${}^{\dagger}$ & Input edits: masked code & Masked-operator predictions reflect code edits beyond frequency effects. \\
\citet{hooda2024cacp} & Input counterfactuals: predicates & Counterfactual code edits expose inconsistent handling of control/data flow and names. \\
\citet{li2024projectbias} & Token attribution: project bias & Regularization improves cross-project generalization and adversarial robustness. \\
\citet{wang2024naming} & Input edits: random/misleading names & Both harm code search and clone detection; misleading names are often worse. \\
\citet{orvalho2025mutations}${}^{\dagger}$ & Input edits: equivalent programs & Mutations degrade output prediction; correct answers can accompany flawed explanations. \\
\citet{liang2026tests}${}^{\dagger}$ & Input edits: test-specified behavior & Test-specified rule switching is incomplete, even for models that implement both rules when explicitly instructed. \\
\addlinespace[5pt]

\multicolumn{3}{@{}l}{\textbf{Programming concepts in internal representations}} \\*
\addlinespace[2pt]
\citet{karmakar2021probing} & Probes: code properties & English-pretrained BERT is competitive with code-pretrained models on some probes. \\
\citet{karmakar2024inspect} & Probes: INSPECT & Fifteen probes; structure-aware models often score higher, with task-dependent rankings. \\
\citet{wan2022structural} & Probes/attention: program structure & Hidden states and attention encode recoverable syntactic relations. \\
\citet{hernandezlopez2022astprobe} & Structural probes: ASTs & Full ASTs recoverable from low-dimensional subspaces; middle layers most informative. \\
\citet{troshin2022probing} & Probes: syntax, names, semantics & Syntax and naming properties are decodable; semantic equivalence is harder. \\
\citet{ma2024syntaxsemantics} & Probes: syntax and flow graphs & Syntax broadly recoverable; control/data dependencies vary across models. \\
\citet{sharma2022attention} & Attention/pooling: identifiers & Attention favors identifiers and separators; identifier pooling improves clone detection. \\
\citet{mohammadkhani2022explaining} & Attention: downstream code tasks & Token-category attention varies across tasks; models can fail on simple programs. \\
\citet{anand2024critical} & Probes: token relations & Syntax--identifier links are poorly recovered; larger tested models often trail smaller ones. \\
\citet{jin2024semantics} & Probes: synthetic execution states & Intermediate states are decodable without supervision on execution traces. \\
\citet{sharma2023redundancy} & Neuron selection: token concepts & Small neuron subsets preserve probe accuracy; selected neurons align with code concepts. \\
\citet{haider2024blackbox} & MLP-key inspection/masking & Masking API-associated keys reduces targeted predictions, with smaller effects on other tokens. \\
\citet{sharma2025latent}${}^{\dagger}$ & Clustering: CoCoA concepts & Lexical, syntactic, and semantic clusters; lexical/subword structure persists in deeper layers. \\
\addlinespace[5pt]

\multicolumn{3}{@{}l}{\textbf{Shared representations across programming and natural languages}} \\*
\addlinespace[2pt]
\citet{kargaran2025shared} & Decoding/ablation: shared neurons & Intermediate states and neurons share programming-language and English information. \\
\citet{north2025beyond} & Probes/attention: variable types & Shared attention heads convey type information across three programming languages. \\
\citet{yin2026neuronguided} & Probes, anonymization, neuron ablation & Language-specific neurons coexist with more invariant middle-layer representations. \\
\citet{naganna2026forminvariant}${}^{\dagger}$ & Patching: arithmetic across formats & Shared neurons across symbols, text, and Python; cross-format patches rescue selected errors. \\
\addlinespace[5pt]

\multicolumn{3}{@{}l}{\textbf{Causal and mechanistic analyses}} \\*
\addlinespace[2pt]
\citet{heimersheim2023docstrings}${}^{\dagger}$ & Circuit tracing: Python docstrings & Compositional argument-copying circuit explains part of a small attention-only model's behavior. \\
\citet{conmy2023circuits} & Automated edge selection: ACDC & Recovers task-relevant subgraphs, including the docstring-copying setting. \\
\citet{templeton2024scaling}${}^{\dagger}$ & SAE clamping: code features & Feature clamping induces insecure code, backdoors, or false bug reports in selected examples. \\
\citet{lucchetti2025types} & Activation steering: type prediction & Repairs induced failures; middle-layer effects and Python--TypeScript transfer vary by edit. \\
\citet{rai2025interference} & Circuit analysis: parentheses & Faulty components override sound computations on synthetic parentheses tasks. \\
\citet{tahimic2025codecorrectness}${}^{\dagger}$ & SAE readouts/steering: correctness & Separate features predict errors and steer generation; steering can repair failures or corrupt correct outputs. \\
\citet{he2026codecircuit}${}^{\dagger}$ & Attribution graphs/intervention & Graphs predict correctness; a binary-search case demonstrates causal repair. \\
\citet{abdelsalam2026conflicts}${}^{\dagger}$ & Patching: misleading names/comments & Traces cue-induced errors through selected residual-stream states in curated conflict cases. \\
\citet{chen2026brewing}${}^{\dagger}$ & Probes/context-stripped decoding & In short Python tasks, probe recovery can precede context-stripped decoding; later layers can lose correct answers. \\
\citet{huang2026deepguard} & Probes/training/steering: security & Intermediate-layer vulnerability signals support improved secure generation with combined interventions. \\
\addlinespace[5pt]

\multicolumn{3}{@{}l}{\textbf{Explanation methods and human comparison}} \\*
\addlinespace[2pt]
\citet{chen2024decix} & Input perturbations: DeciX & Explanation accuracy, compactness, and stability improve over evaluated baselines. \\
\citet{paltenghi2024followup} & Attention/gaze: code exploration & Processed follow-up attention captures aspects of developers' code-reading transitions. \\
\end{longtable}
\endgroup

\clearpage
\section{Selecting and generating negative examples}
\label{app:negative-strategies}

\begingroup
\setlength{\intextsep}{6pt}
\setlength{\abovecaptionskip}{5pt}
\setlength{\belowcaptionskip}{0pt}
\begin{figure}[!ht]
\centering
\includegraphics[width=\linewidth]{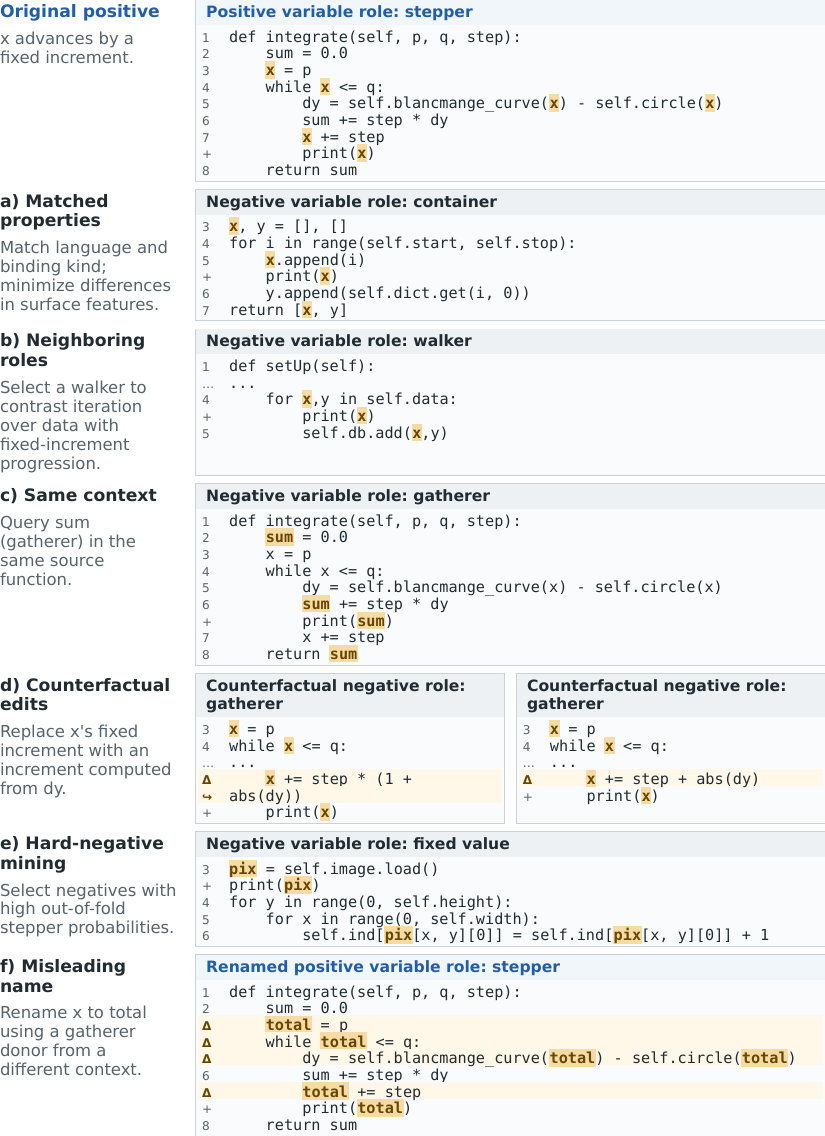}
\caption{\textbf{Strategies for selecting or creating negative examples, illustrated for the stepper role.}
Bold gold marks targets; headers give their roles.
Selected negatives lack the target annotation, counterfactuals change the role, and renaming preserves it.
Illustrative \texttt{print} readouts ($+$) immediately follow the statement or loop header containing the occurrence of interest, making that text available to causal attention.
$\Delta$ marks edits, ellipses omit lines, and arrows mark wrapping.
Examples come from annotated Python training data.}
\label{fig:negative-strategies-stepper}
\end{figure}
\clearpage
\begin{figure}[!ht]
\centering
\includegraphics[width=\linewidth]{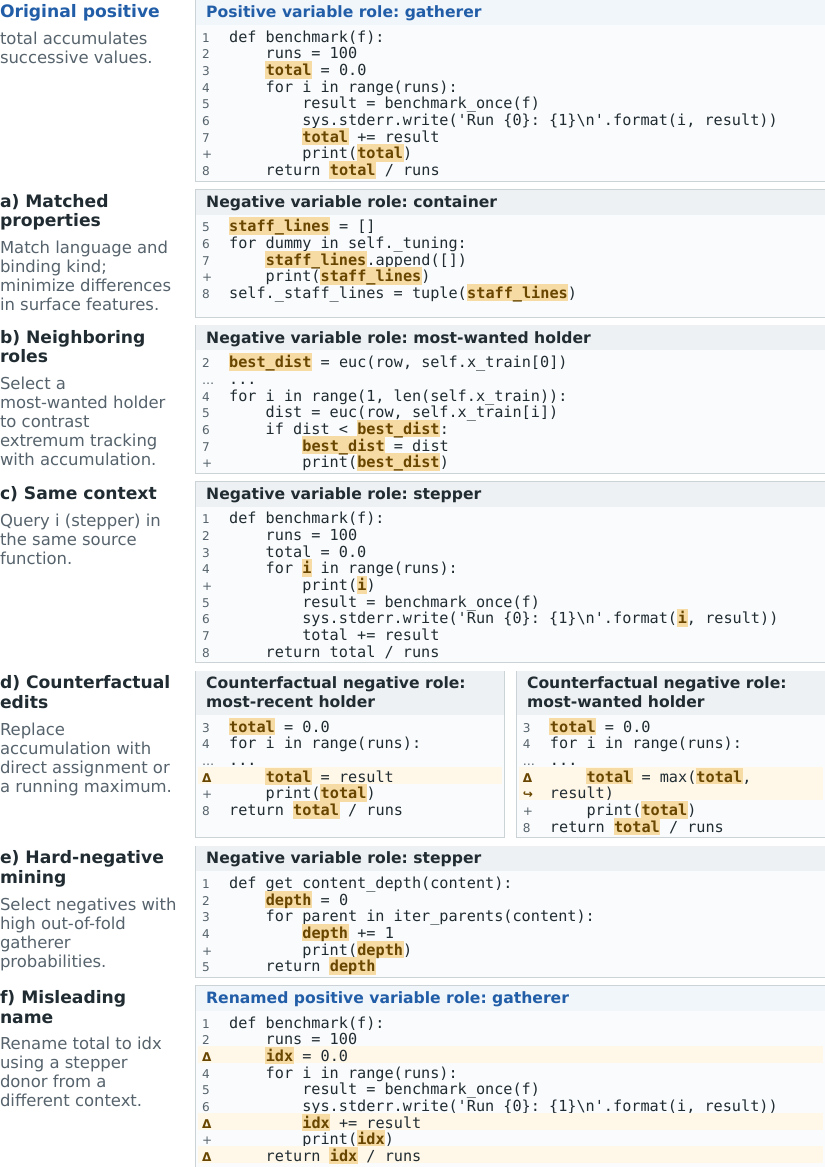}
\caption{\textbf{Strategies for selecting or creating negative examples, illustrated for the gatherer role.}
The six strategies vary the queried binding, context, update, or name.
Renaming preserves the role and all non-target code; donor and recipient role sets are disjoint and their source groups differ.
The miner uses target-masked code and logistic regression in three source-group-held-out folds within training.
Added \texttt{print} readouts, highlighting and edit markers follow Figure~\ref{fig:negative-strategies-stepper}.}
\label{fig:negative-strategies-gatherer}
\end{figure}
\clearpage
\endgroup

\clearpage
\section{Different realizations of variable roles}
\label{app:role-realizations}

\lstset{
    basicstyle=\ttfamily\scriptsize,
    columns=fullflexible,
    keepspaces=true,
    showstringspaces=false,
    frame=single,
    framerule=0.3pt,
    xleftmargin=2pt,
    xrightmargin=2pt,
    aboveskip=2pt,
    belowskip=2pt,
    breaklines=true,
    tabsize=2,
}

\begin{figure*}[t]
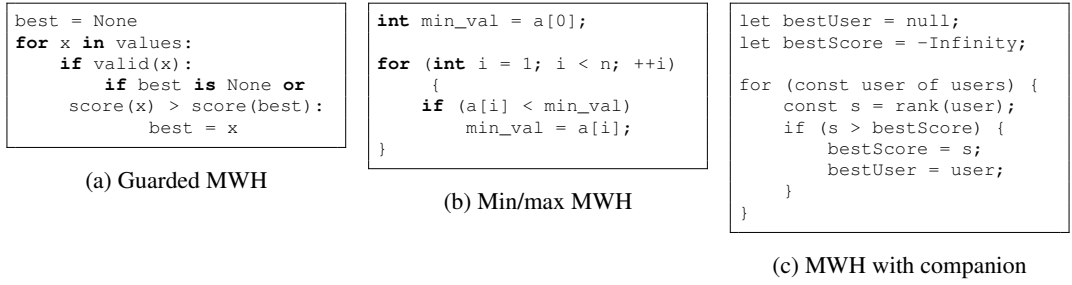

\centering

\begin{subfigure}[t]{0.315\textwidth}
\begin{lstlisting}[language=Python]
best = None
for x in values:
    if valid(x):
        if best is None or score(x) > score(best):
            best = x
\end{lstlisting}
\caption{Guarded MWH}
\end{subfigure}
\hfill
\begin{subfigure}[t]{0.315\textwidth}
\begin{lstlisting}[language=C]
int min_val = a[0];

for (int i = 1; i < n; ++i) {
    if (a[i] < min_val)
        min_val = a[i];
}
\end{lstlisting}
\caption{Min/max MWH}
\end{subfigure}
\hfill
\begin{subfigure}[t]{0.315\textwidth}
\begin{lstlisting}
let bestUser = null;
let bestScore = -Infinity;

for (const user of users) {
    const s = rank(user);
    if (s > bestScore) {
        bestScore = s;
        bestUser = user;
    }
}
\end{lstlisting}
\caption{MWH with companion}
\end{subfigure}

\caption{\textbf{Different realizations of the most-wanted holder role.}
A guarded MWH updates only when an additional eligibility condition holds;
a min/max MWH retains the extremal value observed so far;
an MWH with a companion variable stores the selected object while another variable tracks the selection criterion.}
\label{fig:mwh_realizations}
\end{figure*}

\begin{figure}[t]
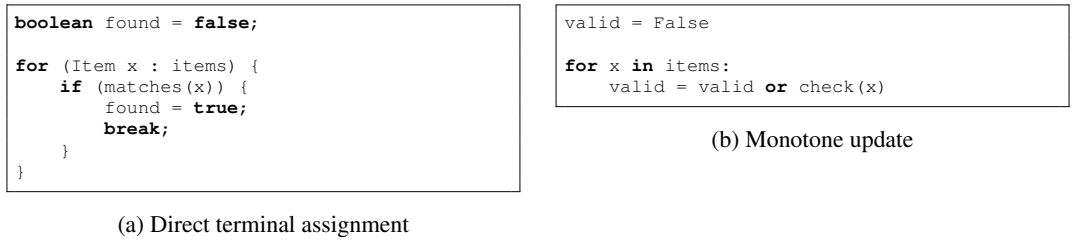

\centering

\begin{subfigure}[t]{0.48\linewidth}
\begin{lstlisting}[language=Java]
boolean found = false;

for (Item x : items) {
    if (matches(x)) {
        found = true;
        break;
    }
}
\end{lstlisting}
\caption{Direct terminal assignment}
\end{subfigure}
\hfill
\begin{subfigure}[t]{0.48\linewidth}
\begin{lstlisting}[language=Python]
valid = False

for x in items:
    valid = valid or check(x)
\end{lstlisting}
\caption{Monotone update}
\end{subfigure}

\caption{\textbf{Different realizations of the one-way flag role.}
The direct form uses a terminal assignment, typically inside a branch.
The monotone form repeatedly combines the previous state with a new condition.
Both allow state changes in only one direction.}
\label{fig:owf_realizations}
\end{figure}

\begin{figure}[t]
\centering

\begin{subfigure}[t]{0.48\linewidth}
\begin{lstlisting}[language=C++]
for (int i = 0; i < n; ++i) {
    process(data[i]);
}
\end{lstlisting}
\caption{Numeric loop iterator}
\end{subfigure}
\hfill
\begin{subfigure}[t]{0.48\linewidth}
\begin{lstlisting}[language=Python]
count = 0
for elt in arr:
    count += 1
\end{lstlisting}
\caption{Predictive accumulator}
\end{subfigure}

\caption{\textbf{Different realizations of the stepper role.}
A numeric iterator follows a regular progression controlling loop iteration.
A predictive accumulator advances by a regular step to represent the position or state expected at the next iteration.}
\label{fig:stepper_realizations}
\end{figure}

\begin{figure}[t]
\centering

\begin{subfigure}[t]{0.48\linewidth}
\begin{lstlisting}[language=Python]
for item in records:
    consume(item)
\end{lstlisting}
\caption{Implicit iterator}
\end{subfigure}
\hfill
\begin{subfigure}[t]{0.48\linewidth}
\begin{lstlisting}[language=C]
Node *node = head;

while (node != NULL) {
    visit(node);
    node = node->next;
}
\end{lstlisting}
\caption{Structural iterator}
\end{subfigure}

\caption{\textbf{Different realizations of the walker role.}
An implicit iterator takes successive elements exposed by the loop construct.
A structural iterator follows explicit links in the data structure.}
\label{fig:walker_realizations}
\end{figure}

\begin{figure}[t]
\centering

\begin{subfigure}[t]{0.48\linewidth}
\begin{lstlisting}
let current = start;
let previous = null;

while (current) {
    current = current.next;
    previous = current;
}
\end{lstlisting}
\caption{Assigned after master change}
\end{subfigure}
\hfill
\begin{subfigure}[t]{0.48\linewidth}
\begin{lstlisting}[language=C++]
Node* current = head;
Node* previous = nullptr;

while (current) {
    previous = current;
    current = current->next;
}
\end{lstlisting}
\caption{Assigned before master change}
\end{subfigure}

\caption{\textbf{Different realizations of the follower role.}
A follower derives its value from a changing master variable.
Assignment order determines whether it captures the master's value before or after a transition.}
\label{fig:follower_realizations}
\end{figure}

\begin{figure}[t]
\centering

\begin{subfigure}[t]{0.48\linewidth}
\begin{lstlisting}[language=Ruby]
last_member = nil

group.each do |member|
  last_member = member
end
\end{lstlisting}
\caption{Member holder}
\end{subfigure}
\hfill
\begin{subfigure}[t]{0.48\linewidth}
\begin{lstlisting}[language=Python]
latest = None

for event in events:
    if event.accepted:
        latest = event
\end{lstlisting}
\caption{Loop-updated holder}
\end{subfigure}

\caption{\textbf{Different realizations of the most-recent holder role.}
A member holder receives successive collection elements.
A loop-updated holder records the latest value satisfying an update condition.
The stored value is replaced at every element or only at selected events, respectively.}
\label{fig:mrh_realizations}
\end{figure}

\clearpage
\section{Matched probe and SAE steering}
\label{app:matched-steering}

We compare probe and SAE directions under the same intervention budget, alongside random residual and random SAE directions.
The exploratory panel covers stepper, gatherer, most-wanted holder, and walker in Gemma~2~2B (layer~13), Gemma~3~12B (layer~24), and Llama~3.1~8B (layer~16).
Each role has twelve calibration tasks and twelve Test tasks, balanced by role presence in the reference solution and shared across models.
These are intact batches selected from the original MBPP cohort without consulting generation outcomes.
Eleven model--role cells pass the generation and evaluation checks; Llama walker is excluded because its zero-dose output differs from the unsteered baseline.

\paragraph{A common budget and calibration criterion.}
We add each normalized direction at generated-token residual positions in one layer, with magnitudes of 1\%, 3\%, or 10\% of the median calibration residual norm.
The zero-strength option is also eligible.
Probe and SAE share the strength maximizing mean calibration induction accompanied by passing tests, subject to each losing at most five percentage points of correctness; ties favor the smaller strength.
The selected strength is fixed for Test and applied to both random controls.
With twelve calibration tasks, the correctness constraint permits no net loss of a passing program.
All conditions retain the same four-task batches, prompts, and greedy 256-token budget, and the zero-dose control must reproduce baseline tokens exactly.
The completed comparison contains 3,696 evaluated generations.

\begin{figure}[!ht]
\centering
\includegraphics[width=\linewidth]{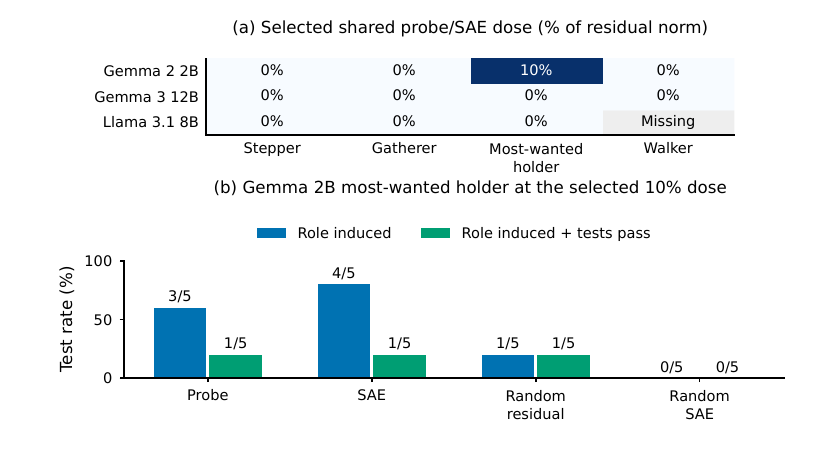}
\caption{Correctness-constrained calibration usually selects no intervention.
(a) Ten of eleven completed cells select zero; the gray cell is unavailable after a failed zero-dose check.
(b) In the sole nonzero cell, probe and SAE directions induce most-wanted holder on three and four of five eligible Test tasks, respectively, but each produces only one induction that passes the supplied tests.
Bars include random controls on the same tasks; fractions show the small conditional denominators.}
\label{fig:matched-steering}
\end{figure}

\paragraph{Role induction often accompanies test failure.}
Gemma~2~2B most-wanted holder selects the 10\% dose (Figure~\ref{fig:matched-steering}).
Its probe and SAE induction rates are 3/5 and 4/5, compared with 1/5 for the random residual direction and 0/5 for the random SAE direction.
Induction accompanied by passing tests is 1/5 for the probe, SAE, and random residual direction, and 0/5 for the random SAE direction.
Overall Test correctness remains 4/12 for both target directions and falls to 3/12 for both controls.
The remaining ten cells select zero because nonzero strengths provide no better feasible calibration objective.
This result distinguishes the ability to alter role annotations from reliable control under a correctness constraint.
These estimates concern positive steering at one layer in small, overlapping task cohorts conditioned on baseline role presence.

\clearpage
\section{Activation contexts of role-associated SAE features}
\label{app:feature-contexts}

The displays below cover all eleven roles with validation-selected single-feature predictors from Gemma~3~4B layers~9 and~17.
Role colors and symbols follow Figure~\ref{fig:taxonomy}.
Each row gives a Neuronpedia description, corpus code (left and ``Also''), and public prose (right).
Highlight intensity is linear from zero to the displayed bound, fixed per latent; compare magnitudes within a latent.
``Lower predicts role'' identifies negative polarity in the original probe selection, so strong highlighting need not occur on the target variable.

\begin{center}
\includegraphics[width=\linewidth]{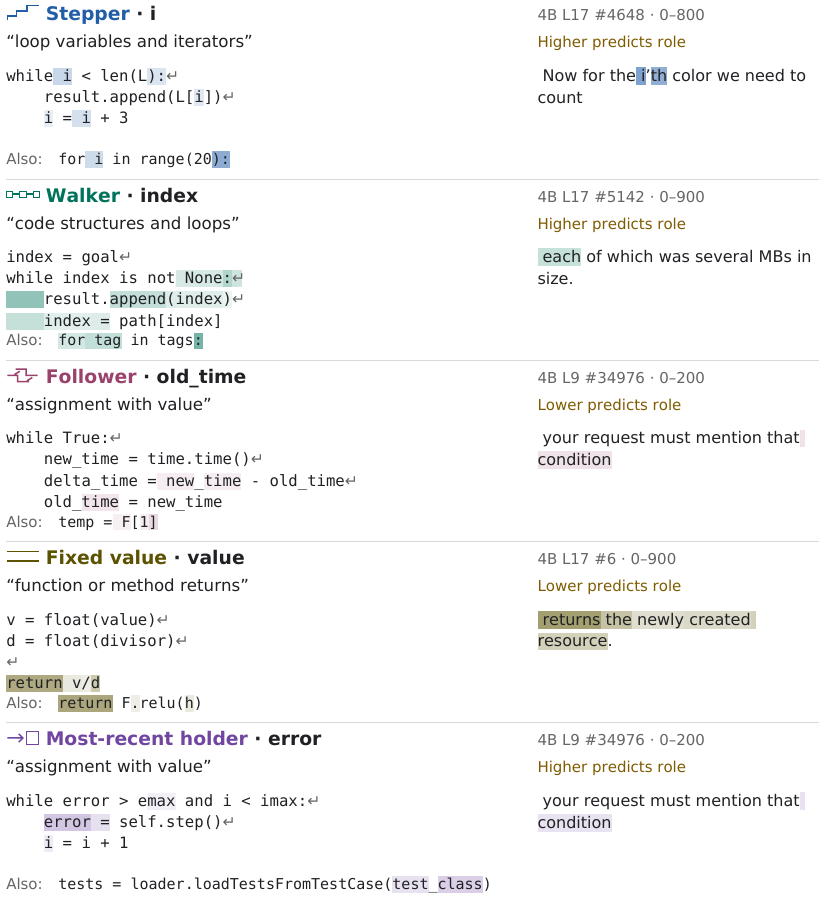}
\captionof{figure}{\textbf{Selected features for progression, traversal, retention, and fixed values.}
Row titles name the role and corpus variable; 4B denotes Gemma~3~4B.
Follower and most-recent holder share feature~34976 with opposite predictive signs.
Return marks denote newline tokens, and ``Also'' introduces a separate corpus context.}
\label{fig:role-latents-first}
\end{center}

All ten distinct latents were measured on 41 hand-selected training snippets.
Code crops retain their standalone-input activations; public windows retain their original Neuronpedia contexts.
These examples illustrate activity, not prevalence.

\clearpage
\section*{Descriptions and activation contexts (continued)}

\begin{center}
\includegraphics[width=\linewidth]{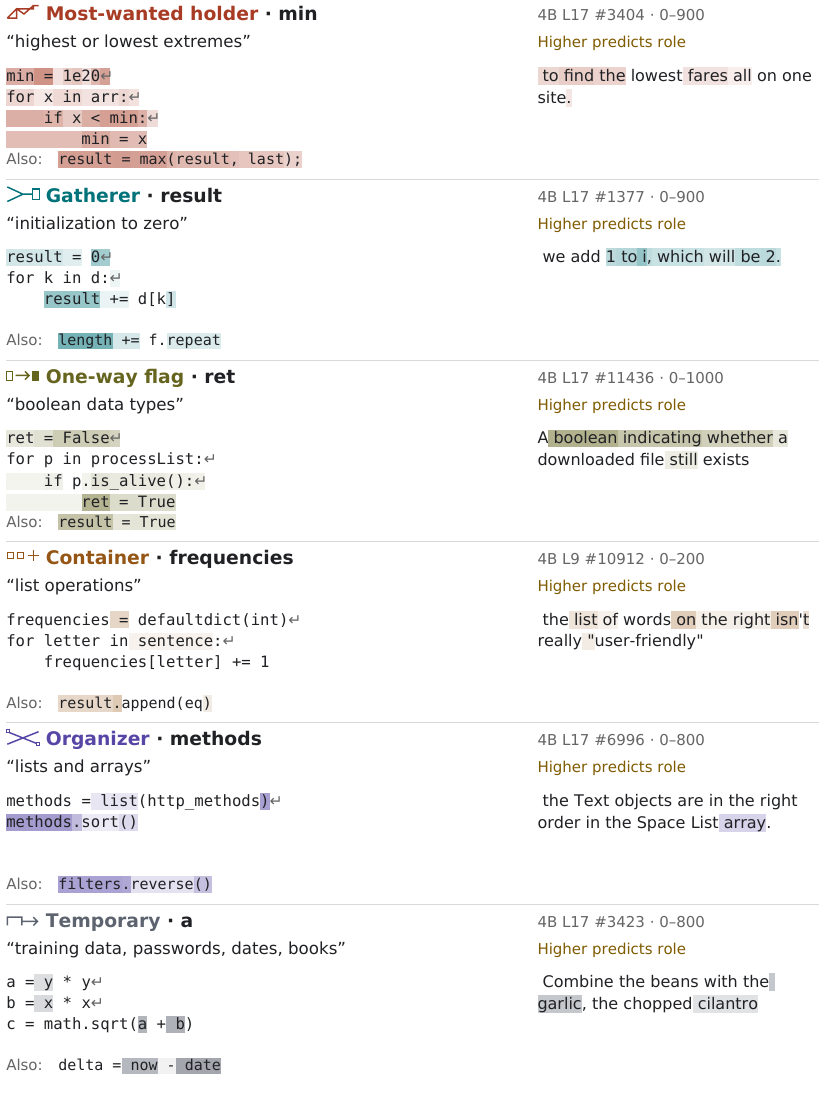}
\captionof{figure}{\textbf{Extrema, accumulation, flags, collections, and intermediate computations.}
The displays continue Figure~\ref{fig:role-latents-first} with the same conventions and per-latent scales.
Descriptions can name broad operations or disparate contexts, rather than the variable role itself.}
\label{fig:role-latents-second}
\end{center}

Matched and reviewed-counterfactual AUROCs are 0.848/0.474 for most-wanted holder, 0.949/0.354 for one-way flag, and 0.764/0.536 for stepper.
Contextual activity alone does not establish role specificity.

\clearpage
\section{Qualitative outcomes of SAE steering}
\label{app:steering-examples}

The larger Gemma~2~2B and Gemma~3~4B experiments from Section~\ref{sec:causal-influence} also permit inspection of complete generated programs.
Across 292 evaluated target--task pairs, eight failing-to-passing transitions correspond to six distinct model--task pairs.
They include most-frequent-item search, counting sort, largest row sum, cuboid volume, product divided by length, and maximum subarray sum.
We show selected repairs and failures alongside an unfiltered seeded random sample in Figure~\ref{fig:steering-examples}.
All displayed test outcomes were reproduced on the complete programs using the three supplied assertions.

The selected repairs can reveal useful changes, such as replacing a stub with an explicit search, without establishing that the targeted role caused the improvement.
In the broader gallery, cuboid and product repairs pass all three tests while retaining a zero target-role annotation.
Conversely, an unchanged correct program may already implement the role semantically even when its original annotation is absent.
The six random cases drawn across the full gallery all fail the supplied tests; the draw was not repeated to obtain favorable examples.
These observations motivate reading role annotations, complete code, and correctness together.

\clearpage
\section{Consolidated quantitative analysis of causal interventions}
\label{app:causal-consolidated}

We connect the selected-feature results in Figure~\ref{fig:causal_scores_and_pass_at_1} and the structural changes in Table~\ref{tab:causal-role-syntax} to all tested SAE features and the normalized probe/SAE experiments.
The two displays summarize complementary measurements from the same Gemma~2~2B and Gemma~3~4B causal runs.
The analysis adds the completed Gemma~3~12B feature-scale run, a five-fit Gemma~3~12B stepper pilot, and the eleven model--role cells in Appendix~\ref{app:matched-steering}.

\begin{figure}[!ht]
\centering
\includegraphics[width=\linewidth]{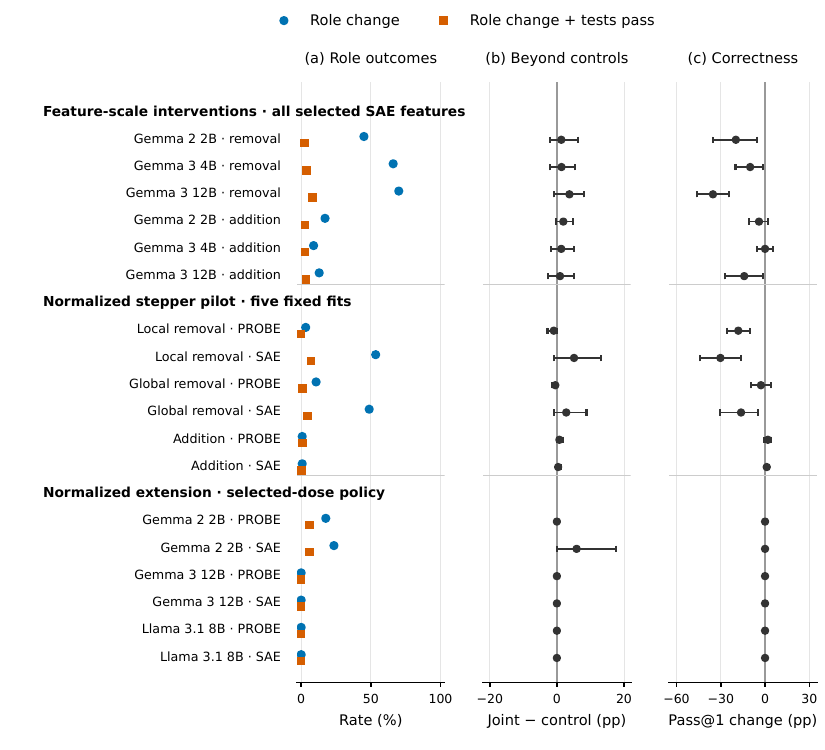}
\caption{\textbf{Annotation changes substantially exceed changes accompanied by passing tests.}
(a) Role disappearance for removal and induction for addition, with the joint passing outcome on the same baseline-defined cohort.
(b) Joint outcome minus the matched random-control mean: random residual directions for probes and random SAE directions for SAE targets.
(c) Intervention-minus-baseline correctness on that cohort.
Rates weight target--task enrollments within each protocol; 95\% intervals resample unique tasks 4,000 times, retaining all associated features, fits, and controls together.
Normalized addition averages include selected zero doses; the stepper pilot uses Gemma~3~12B.
The three protocol groups use different cohorts and perturbation scales.}
\label{fig:causal-consolidated}
\end{figure}

Across all historical features, removal changes 41/91, 72/109, and 131/187 target--task outcomes for Gemma~2~2B, Gemma~3~4B, and Gemma~3~12B, respectively.
Only 2/91, 4/109, and 15/187 also pass the supplied tests; requiring both outputs to pass leaves 0, 3, and 13 outcomes.
The joint outcome exceeds matched random controls by only 1.3, 1.4, and 3.7 percentage points for removal, and 1.9, 1.3, and 0.9 points for addition; all six corresponding intervals include zero.
These enrollment counts include repeated tasks across features and are not independent sample sizes.
The removal cohorts contain 32, 32, and 63 unique tasks, respectively; reference-positive addition uses 53, 49, and 45 unique Test tasks.

\clearpage
\subsection{Comparison units, calibration, and structural survival}

\paragraph{Protocol-specific comparisons.}
We test eleven SAE directions in Gemma~2~2B and Gemma~3~4B and nine in Gemma~3~12B, covering five roles.
They share a union of 121 MBPP tasks with role-specific enrollment; removal uses both splits, while induction uses Test only.
Six positive multipliers are calibrated for annotation induction, with ten same-layer random SAE controls per target.
The baseline passing counts are 33/121, 14/121, and 67/121.
The archived feature scales are not normalized across models, and twenty of the shared package's twenty-two scales differ from the corresponding selection artifact's training-positive maximum.
They are treated as executed settings with unresolved scale provenance.
Consequently, these results describe the recorded interventions without attributing cross-model differences to model size or a common dose.

The normalized stepper pilot uses Gemma~3~12B layer~31, five fitted probes, and the same SAE feature in every fit.
It has 48 calibration and 144 Test tasks, including 43 eligible removal and 53 induction tasks.
Four fits select zero and one selects a 1\% residual-norm dose under the correctness constraint.
In that nonzero fit, probe and SAE directions each induce stepper on 2/53 tasks; two probe outputs and one SAE output pass, including one probe change that preserves baseline correctness.
Repeated SAE outcomes and overlapping probe training sets are not independent replications.
Local projection affects one generated-token layer; global projection also alters residual writers and prompt processing.
The smaller extension in Appendix~\ref{app:matched-steering} instead covers four roles in twelve-task splits and selects zero in ten of eleven completed cells.

\begin{figure}[!ht]
\centering
\includegraphics[width=\linewidth]{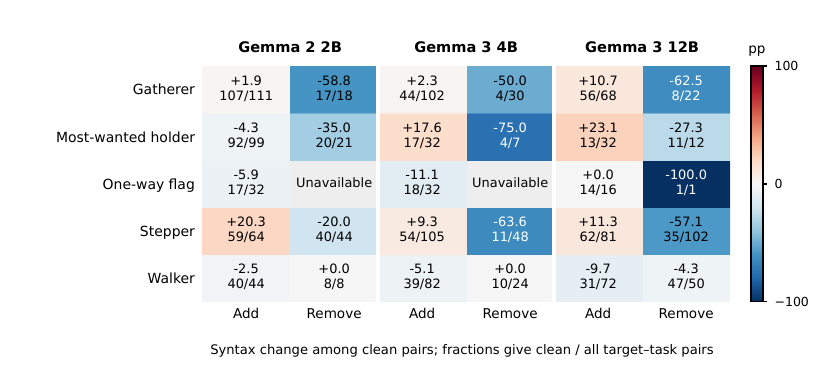}
\caption{\textbf{Large structural shifts can describe a small surviving subset.}
Values reproduce Table~\ref{tab:causal-role-syntax} and extend it to Gemma~12B: syntax-frequency changes in percentage points among pairs with two clean Tree-sitter parses.
Fractions show retained clean pairs over all target--task pairs.
Markers follow Table~\ref{tab:causal-role-syntax}; addition combines Test sufficiency and reference-negative cohorts.
Gray cells have no eligible removal cohort.}
\label{fig:causal-syntax-survival}
\end{figure}

\paragraph{Syntax and correctness condition different subsets.}
For Gemma~3~4B removal, gatherer syntax uses 4/30 pairs from two distinct tasks, and stepper syntax uses 11/48 pairs from five tasks.
The reported shifts therefore describe parseable survivors, not all interventions.
Among the three clean stepper pairs where both outputs pass, the \texttt{range}-call frequency does not change.

\paragraph{Scope of inference.}
Within each protocol, target and control outcomes use the same eligible tasks and retained denominator.
The joint metric requires the intervention output to pass; preserving an already correct program additionally requires baseline correctness.
Intervals condition on selected features, fitted probes, calibration, and sampled controls; all-zero samples can yield degenerate intervals.
The evaluated records share twelve task IDs between historical runs and the smaller extension, four between the pilot and extension, and none between historical runs and the pilot.
No effect is pooled across protocols.
Finite, prompt-visible tests, generated-code truncation, and missing historical zero-dose/batch-parity audits limit claims of semantic control.

\end{document}